\def\year{2022}\relax
\documentclass[letterpaper]{article} 
\usepackage{aaai22}  
\usepackage{times}  
\usepackage{helvet}  
\usepackage{courier}  
\usepackage[hyphens]{url}  
\usepackage{graphicx} 
\usepackage{natbib}  
\usepackage{caption} 
\DeclareCaptionStyle{ruled}{labelfont=normalfont,labelsep=colon,strut=off} 
\usepackage{algorithm}
\usepackage{algorithmic}

\usepackage{newfloat}
\usepackage{listings}
\floatstyle{ruled}
\newfloat{listing}{tb}{lst}{}
\floatname{listing}{Listing}
\usepackage[utf8]{inputenc}
\usepackage[algo2e]{algorithm2e}
\usepackage{multirow}
\usepackage{graphicx}
\usepackage{subcaption}
\usepackage{caption}
\usepackage{placeins}
\usepackage{booktabs}
\usepackage{color}
\usepackage{amsmath}
\usepackage{amsfonts}
\usepackage{amssymb}
\usepackage{amsthm}
\usepackage{enumitem}

\usepackage{lipsum}

\theoremstyle{plain}
\newtheorem{theorem}{Theorem}

\theoremstyle{definition}
\newtheorem{definition}[theorem]{Definition}
\theoremstyle{corollary}
\newtheorem{corollary}[theorem]{Corollary}
\theoremstyle{proposition}
\newtheorem{proposition}[theorem]{Proposition}

\theoremstyle{remark}

\newcommand{\RNum}[1]{\uppercase\expandafter{\romannumeral #1\relax}}
\newcommand{\bm}[1]{\boldsymbol{#1}}

\newcommand{\Black}[1]{\textcolor[rgb]{0.00,0.00,0.00}{#1}}

\newcommand{\reviseyq}[1]{\Black{#1}}
\newcommand{\revisexin}[1]{\Black{#1}}

\title{Graph Convolutional Networks with Dual Message Passing for \\
Subgraph Isomorphism Counting and Matching}

\author {
    Xin Liu,
    Yangqiu Song
}
\affiliations {
    Department of CSE, the Hong Kong University of Science and Technology \\
    \{xliucr, yqsong\}@cse.ust.hk
}

\begin{document}
\maketitle
\begin{abstract}
Graph neural networks (GNNs) and message passing neural networks (MPNNs) have been proven to be expressive for subgraph structures in many applications. 
Some applications in heterogeneous graphs require explicit edge modeling, such as subgraph isomorphism counting and matching.
However, existing message passing mechanisms are not designed well in theory.
In this paper, we start from a particular edge-to-vertex transform and exploit the isomorphism property in the edge-to-vertex dual graphs.
We prove that searching isomorphisms on the original graph is equivalent to searching on its dual graph.
Based on this observation, we propose dual message passing neural networks (DMPNNs) to enhance the substructure representation learning in an asynchronous way for subgraph isomorphism counting and matching as well as unsupervised node classification.
Extensive experiments demonstrate the robust performance of DMPNNs by combining both node and edge representation learning in synthetic and real heterogeneous graphs. \revisexin{
Code is available at \url{https://github.com/HKUST-KnowComp/DualMessagePassing}.}
\end{abstract}

\section{Introduction}

Graphs have been widely used in various applications across domains from chemoinformatics to social networks.
The isomorphism is one of the important properties in graphs, and analysis on subgraph isomorphisms is useful in real applications.
For example, \reviseyq{we} can determine the properties of compounds by finding functional group information in chemical molecules~\cite{gilmer2017neural}\revisexin{; some} sub-structures in social networks are regarded as irreplaceable features in recommender systems~\cite{ying2018graph}.
The challenge \reviseyq{of finding} subgraph isomorphisms \revisexin{requires} the exponential \reviseyq{computational} cost.
\revisexin{Particularly, finding and counting require global inference to oversee the whole graph.}
Existing counting and matching algorithms are designed for some query patterns up to a certain size (e.g., 5), and some of them cannot directly apply \reviseyq{to} heterogeneous graphs \reviseyq{where vertices and edges are labeled with types}~\cite{bhattarai2019ceci,sun2020in}.

There has been more attention to using deep learning to count or match subgraph isomorphisms.
\citeauthor{liu2020neural}~(\citeyear{liu2020neural}) \reviseyq{designed} a general end-to-end framework to predict the number of subgraph isomorphisms on heterogeneous graphs, and \citeauthor{ying2020neural}~(\citeyear{ying2020neural}) \reviseyq{combined} node embeddings and voting to match subgraphs.
They \reviseyq{found} that neural networks \revisexin{could} speed up 10 to 1,000 times compared with traditional searching algorithms.
\citeauthor{xu2019how}~(\citeyear{xu2019how}) and \citeauthor{morris2019weisfeiler}~(\citeyear{morris2019weisfeiler}) \reviseyq{showed} that graph neural networks (GNNs) based on message passing are at most as powerful as the WL test~\cite{weisfeiler1968reduction}, and \citeauthor{chen2020can}~(\citeyear{chen2020can}) further \reviseyq{analyzed} the upper-bound of message passing and $k$-WL for subgraph isomorphism counting.
\reviseyq{These studies show that} it is theoretically possible for neural methods to count larger patterns in complex graphs.
In heterogeneous graphs, edges play an important role in checking and searching isomorphisms because graph isomorphisms require taking account of graph adjacency and edge types.
However, existing message passing mechanisms have not paid enough attention to edge representations~\cite{gilmer2017neural,schlichtkrull2018modeling,vashishth2020composition,jin2021power}.

In this paper, we discuss a particular edge-to-vertex transform and find the one-to-one correspondence between subgraph isomorphisms of \reviseyq{original} graphs and subgraph isomorphisms of their \reviseyq{corresponding} edge-to-vertex dual graphs.
This property suggests that searching isomorphisms on the original graph is equivalent to searching on its dual graph.
\reviseyq{Based on this observation and the theoretical guarantee, we} propose \reviseyq{new} dual message passing networks (DMPNNs) to learn node and edge representations \reviseyq{simultaneously} in \reviseyq{the} aligned space.
Empirical results show the effectiveness of DMPNNs on \reviseyq{all} homogenerous \reviseyq{and} heterogeneous graphs, synthetic data or real-life data.

Our main contributions are summarized as follows:
\begin{enumerate}
    \item We prove that there is a one-to-one correspondence between isomorphisms of connected directed heterogeneous multi-graphs with reversed edges and isomorphisms between their edge-to-vertex dual graphs.
    \item We propose dual message passing mechanism and design \reviseyq{the DMPNN model} to explicitly model edges and align node and edge representations in the same space.
    \item \revisexin{We empirically demonstrate that DMPNNs can count subgraph isomorphisms more accurately and match isomorphic nodes more correctly}. DMPNNs also surpass competitive baselines on unsupervised node classification, indicating the necessity of explicit edge modeling \reviseyq{for general graph representation learning}. 
\end{enumerate}
\begin{figure*}[!t]
    \centering
    \begin{subfigure}{.19\textwidth}
        \centering
        \includegraphics[height=4.4cm]{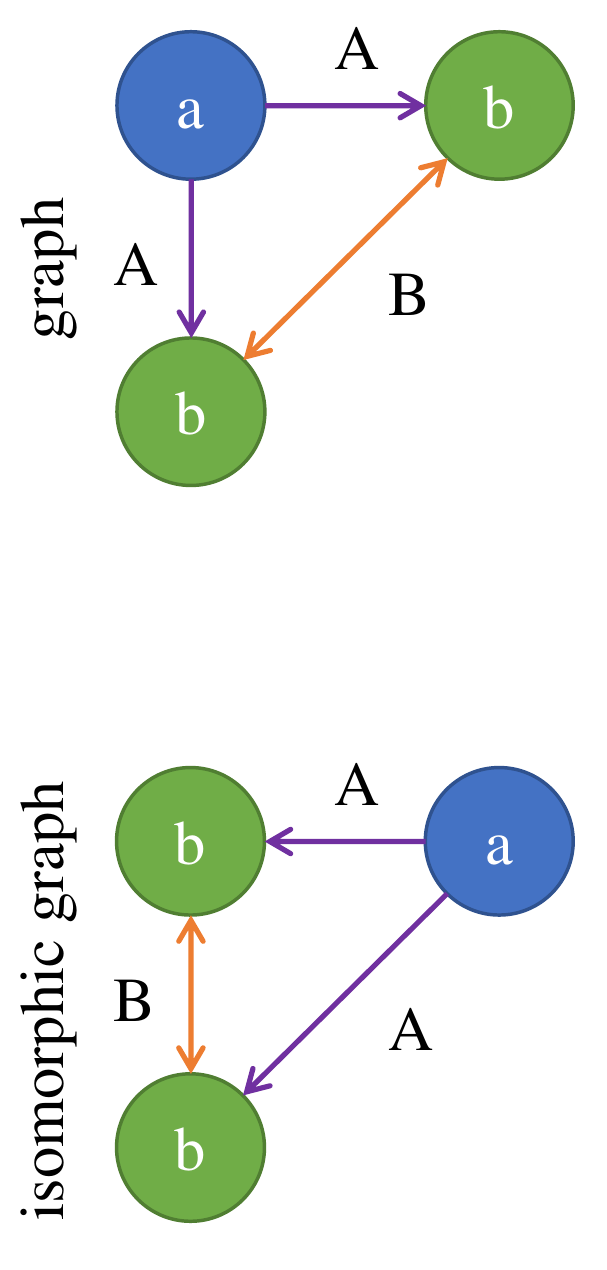}
        \caption{Isomorphism}
        \label{fig:preliminary_iso}
    \end{subfigure}
    \begin{subfigure}{.20\textwidth}
        \centering
        \includegraphics[height=4.4cm]{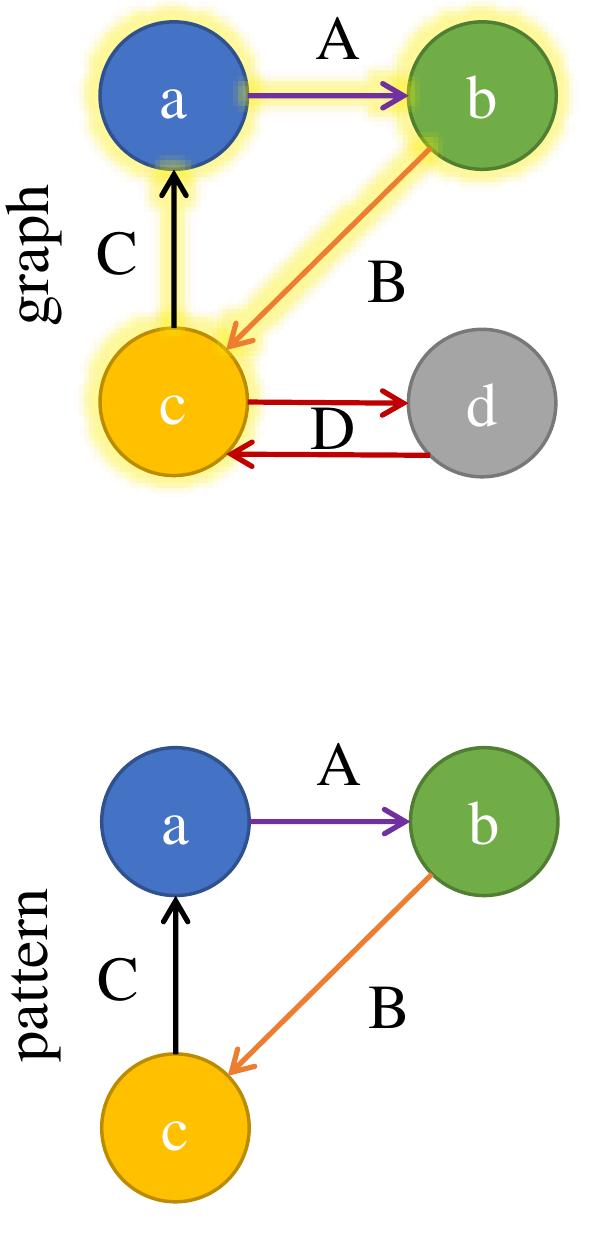}
        \caption{Subgraph isomorphism}
        \label{fig:preliminary_subiso}
    \end{subfigure}
    \begin{subfigure}{.31\textwidth}
        \centering
        \includegraphics[height=4.4cm]{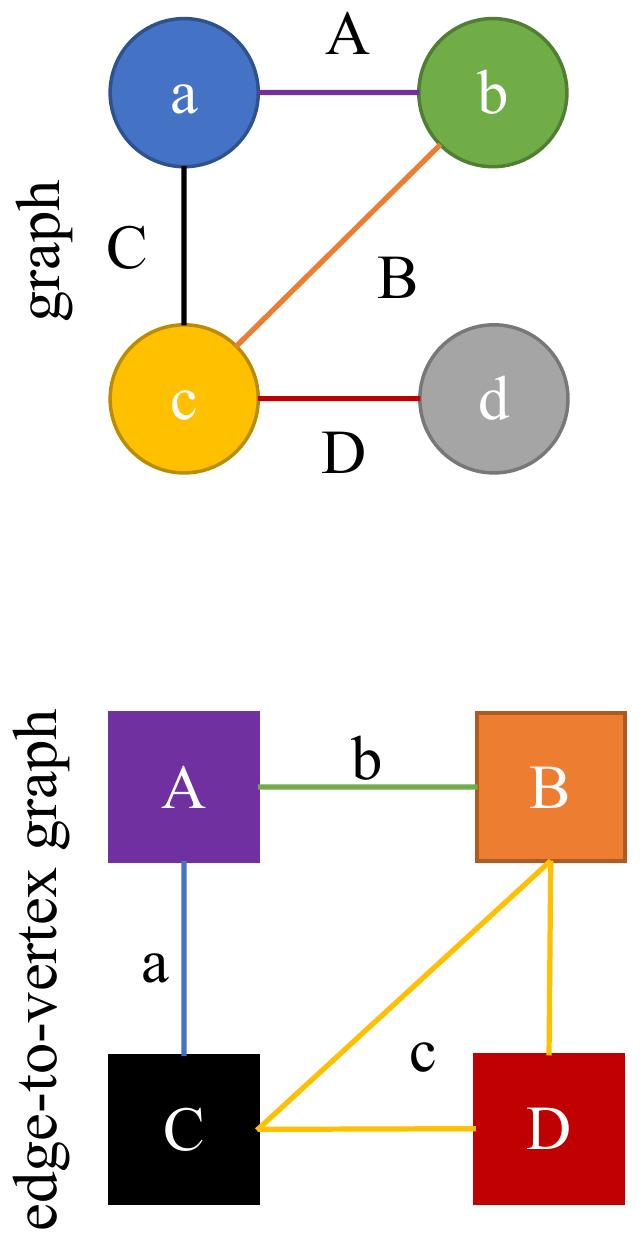}
        \caption{Edge-to-vertex transform (undirected)}
        \label{fig:preliminary_line}
    \end{subfigure}
    \begin{subfigure}{.28\textwidth}
        \centering
        \includegraphics[height=4.4cm]{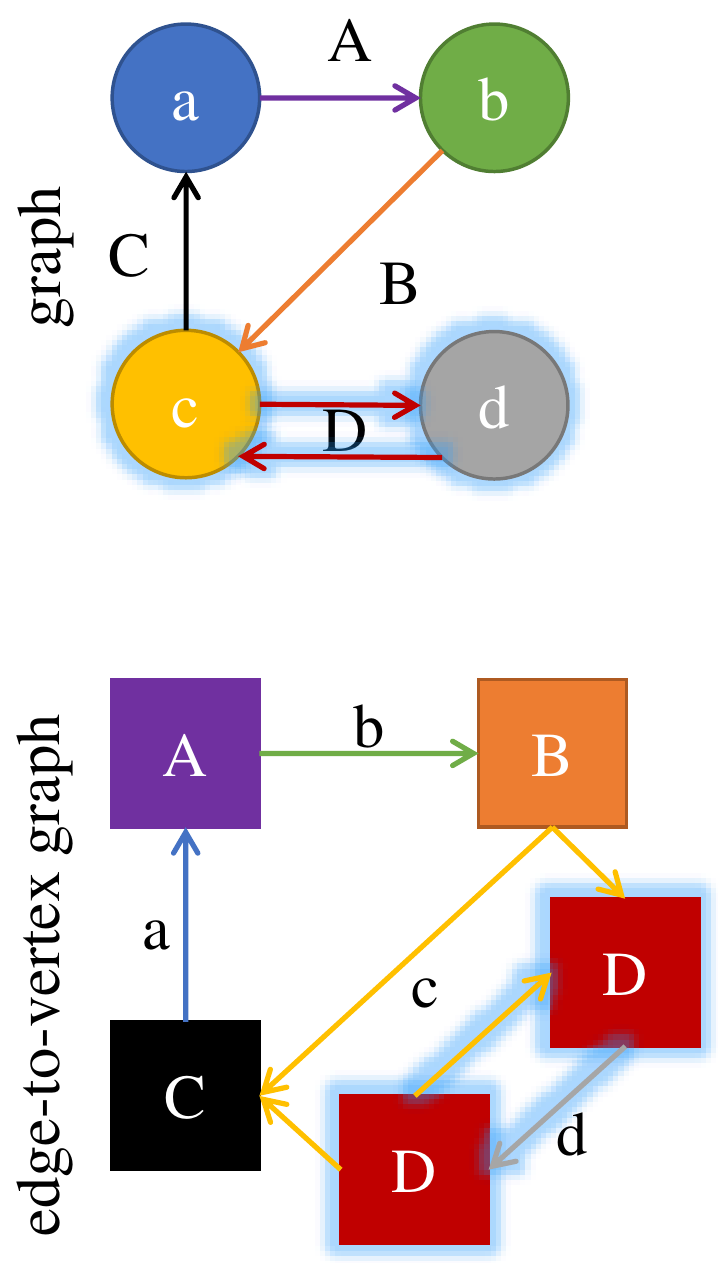}
        \caption{Edge-to-vertex transform (directed)}
        \label{fig:preliminary_line_dir}
    \end{subfigure}
    \vspace{-0.1in}
    \caption{Examples of the isomorphism, subgraph isomorphism, and edge-to-vertex transforms.}
    \vspace{-0.15in}
\end{figure*}

\section{Preliminaries}
To be more general, we assume a graph is a directed heterogeneous multigraph.
Let $\mathcal{G}$ be a graph with a vertex set $\mathcal{V}_\mathcal{G}$ and each vertex with a different \textit{vertex id}, an edge set $\mathcal{E}_\mathcal{G} \subseteq \mathcal{V}_\mathcal{G} \times \mathcal{V}_\mathcal{G}$,
a label function $\mathcal{X}_\mathcal{G}$ that maps a vertex to a \textit{vertex label}, and a label function $\mathcal{Y}_\mathcal{G}$ that maps an edge to a set of \textit{edge labels}.
As we regard each edge can be associated with a set of labels, we can merge multiple edges with the same source and the same target as one edge with multiple labels. 
\revisexin{
A subgraph of $\mathcal{G}$, denoted as $\mathcal{G}_S$, is any graph with $\mathcal{V}_{\mathcal{G}_S} \subseteq \mathcal{V}_\mathcal{G}$, $\mathcal{E}_{\mathcal{G}_S} \subseteq \mathcal{E}_\mathcal{G} \cap (\mathcal{V}_{\mathcal{G}_{S}} \times \mathcal{V}_{\mathcal{G}_{S}})$ satisfying 
$\forall v \in \mathcal{V}_{\mathcal{G}_S}, \mathcal{X}_{\mathcal{G}_S}(v) = \mathcal{X}_{\mathcal{G}}(v)$ and $\forall e \in \mathcal{E}_{\mathcal{G}_S}, \mathcal{Y}_{\mathcal{G}_S}(e) = \mathcal{Y}_{\mathcal{G}}(e)$.
}
To simplify the statement, we let $\mathcal{Y}_\mathcal{G}((u, v)) = \phi$ if $(u, v) \not \in \mathcal{E}_\mathcal{G}$.

\subsection{Isomorphisms and Subgraph Isomorphisms}
\begin{definition}[Isomorphism]
\label{def:iso}
A graph $\mathcal{G}_1$ is \textit{isomorphic} to a graph $\mathcal{G}_2$ if there is a bijection $f: \mathcal{V}_{\mathcal{G}_1} \rightarrow \mathcal{V}_{\mathcal{G}_2}$ such that:
\scalebox{0.950}{\parbox{1.07\linewidth}{
\begin{itemize}
    \item $\forall v \in \mathcal{V}_{\mathcal{G}_1}, \mathcal{X}_{\mathcal{G}_1}(v) = \mathcal{X}_{\mathcal{G}_2}(f(v))$, 
    \item $\forall v' \in \mathcal{V}_{\mathcal{G}_2}, \mathcal{X}_{\mathcal{G}_2}(v') = \mathcal{X}_{\mathcal{G}_1}(f^{-1}(v'))$,
    \item $\forall (u, v) \in \mathcal{E}_{\mathcal{G}_1}, \mathcal{Y}_{\mathcal{G}_1}((u, v)) = \mathcal{Y}_{\mathcal{G}_2}((f(u), f(v)))$,
    \item $\forall (u', v') \in \mathcal{E}_{\mathcal{G}_2}, \mathcal{Y}_{\mathcal{G}_2}((u', v')) = \mathcal{Y}_{\mathcal{G}_1}((f^{-1}(u'), f^{-1}(v')))$.
\end{itemize}
}}
\end{definition}
We write $\mathcal{G}_1 \simeq \mathcal{G}_2$ for such isomorphic property and name $f$ as an \textit{isomorphism}. For example, there are two different isomorphisms between the two triangles in Figure~\ref{fig:preliminary_iso}.
As a special \reviseyq{case}, the isomorphism $f$ between two empty graphs without any vertex is $\{\} \rightarrow \{\}$.

In addition, if a subgraph of $\mathcal{G}_1$ is isomorphic to another graph, then the corresponding bijection function is named as a \textit{subgraph isomorphism}. The formal definition is:
\begin{definition}[Subgraph isomorphism]
\label{def:subiso}
If a subgraph $\mathcal{G}_{1_S}$ of $\mathcal{G}_1$ is isomorphic to a graph $\mathcal{G}_2$ with a bijection $f$,
\revisexin{we say $\mathcal{G}_{1}$ contains a subgraph isomorphic to $\mathcal{G}_2$ and name $f$ as a \textit{subgraph isomorphism}.}
\end{definition}

Subgraph isomorphism related problems commonly refer to two kinds of subgraphs: node-induced subgraphs and edge-induced subgraphs.
In node-induced subgraph related problems, the possible subgraphs require \reviseyq{that for} each vertex in $\mathcal{G}_S$, the associated edges in $\mathcal{G}$ must appear in $\mathcal{G}_S$, i.e., 
\revisexin{
$\mathcal{V}_{\mathcal{G}_S} \subseteq \mathcal{V}_{\mathcal{G}}, \mathcal{E}_{\mathcal{G}_S} = \mathcal{E}_\mathcal{G} \cap (\mathcal{V}_{\mathcal{G}_{S}} \times \mathcal{V}_{\mathcal{G}_{S}})$;}
in edge-induced subgraph related problems, the required subgraphs are restricted by associating vertices that are incident to edges, i.e., \revisexin{$\mathcal{E}_{\mathcal{G}_S} \subseteq \mathcal{E}_{\mathcal{G}}$, $\mathcal{V}_{\mathcal{G}_S} = \{u | (u, v) \in \mathcal{E}_{\mathcal{G}_S}\} \cup \{v | (u, v) \in \mathcal{E}_{\mathcal{G}_S}\}$}. 
Node-induced subgraphs are specific edge-induced subgraphs when $\mathcal{G}$ is connected. 
Hence, we assume all subgraphs mentioned in the following are edge-induced for better generalization. 
Figure~\ref{fig:preliminary_subiso} shows an example of subgraph isomorphism that a graph with four vertices is subgraph isomorphic to the triangle pattern.

\subsection{Edge-to-vertex Transforms}
In graph theory, the line graph of an undirected graph $\mathcal{G}$ is another undirected graph that represents the adjacencies between edges of $\mathcal{G}$, 
e.g., Figure~\ref{fig:preliminary_line}. We extend line graphs to directed heterogeneous multigraphs.
\begin{definition}[Edge-to-vertex transform]
\label{def:line}
A \textit{line graph} (also known as \textit{edge-to-vertex dual graph}) $\mathcal{H}$ of a graph $\mathcal{G}$ is obtained by associating a vertex $v' \in \mathcal{V}_{\mathcal{H}}$ with each edge $e = g^{-1}(v') \in \mathcal{E}_{\mathcal{G}}$ and connecting two vertices $u', v' \in \mathcal{V}_{\mathcal{H}}$ with an edge from $u'$ to $v'$ if and only if the destination of the corresponding edge $d = g^{-1}(u')$ is exact the source of $e = g^{-1}(v')$.
Formally, we have:
\scalebox{0.950}{\parbox{1.07\linewidth}{
\begin{itemize}
    \item $\forall e=(u,v) \in \mathcal{E}_{\mathcal{G}}, 
    \mathcal{Y}_{\mathcal{G}}(e) = \mathcal{X}_{\mathcal{H}}(g(e))$,
    \item $\forall v' \in \mathcal{V}_{\mathcal{H}},
    \mathcal{X}_{\mathcal{H}}(v') = \mathcal{Y}_{\mathcal{G}}(g^{-1}(v'))$,
    \item $\forall d, e \in \mathcal{E}_{\mathcal{G}}, u' = g(d) \in \mathcal{V}_{\mathcal{H}}, v' = g(e) \in \mathcal{V}_{\mathcal{H}}, \\(d.\text{target}=e.\text{source}=v) \rightarrow (\mathcal{Y}_{\mathcal{H}}((u', v')) = \mathcal{X}_{\mathcal{G}}(v)))$,
    \item $\forall e'=(u', v') \in \mathcal{E}_{\mathcal{H}}, d=g^{-1}(u') \in \mathcal{V}_{\mathcal{G}}, e=g^{-1}(v') \in \mathcal{V}_{\mathcal{G}},\\ (d.\text{target} = e.\text{source}) \wedge (\mathcal{Y}_{\mathcal{H}}(e') = \mathcal{X}_{\mathcal{H}}(d.\text{target}))$.
\end{itemize}
}}
\end{definition}
We call the bijection $g: \mathcal{E}_{\mathcal{G}} \rightarrow \mathcal{V}_\mathcal{H}$ as the \textit{edge-to-vertex map}, and write $\mathcal{H}$ as $L(\mathcal{G})$ where $L: \mathcal{G} \rightarrow \mathcal{H}$ \revisexin{corresponds to} the \textit{edge-to-vertex transform}.
There are several differences between undirected line graphs and directed line graphs.
As shown in Figure~\ref{fig:preliminary_line} and Figure~\ref{fig:preliminary_line_dir}, 
\revisexin{except} directions of edges,
\revisexin{an edge with its inverse in the original graph will introduce two corresponding vertices and a pair of reversed edges in between in the line graph.}

There are many properties in the edge-to-vertex graph.
As the vertices of the line graph $\mathcal{H}$ corresponds to the edges of the original graph $\mathcal{G}$, some properties of $\mathcal{G}$ that depend only on adjacency between edges may be preserved as equivalent properties in $\mathcal{H}$ that depend on adjacency between vertices.
For example, an independent set in $\mathcal{H}$ corresponds to a matching (also known as independent edge set) in $\mathcal{G}$.
But the edge-to-vertex transform may lose the information of the original graph.
For example, two different graphs may have the same line graph.
We have one observation that if two graphs are isomorphic, their line graphs are also isomorphic;
nevertheless, the converse is not always correct. We will discuss the isomorphism and the edge-to-vertex transform in the next section. 
\section{Isomorphisms vs. Edge-to-vertex Transforms}

The edge-to-vertex transform can preserve adjacency relevant properties of graphs. In this section, we discuss isomorphisms and the edge-to-vertex transform.
\revisexin{Particularly, we analyze the symmetry of isomorphisms in special situations transforming edges to vertices, and we further extend all graphs into this particular kind of structure for searching.}


\begin{proposition}\label{propo:line_iso}
If two graphs $\mathcal{G}_1$ and $\mathcal{G}_2$ are isomorphic with an isomorphism $f: \mathcal{V}_{\mathcal{G}_1} \rightarrow \mathcal{V}_{\mathcal{G}_2}$, then their line graphs $\mathcal{H}_1$ and $\mathcal{H}_2$ are also isomorphic with an isomorphism $f': \mathcal{V}_{\mathcal{H}_1} \rightarrow \mathcal{V}_{\mathcal{H}_2}$ such that $\forall v \in \mathcal{V}_{{\mathcal{H}_1}}, \mathcal{X}_{\mathcal{H}_1}(v) = \mathcal{X}_{\mathcal{H}_2}(f'(v))$ and $\forall v' \in \mathcal{V}_{{\mathcal{H}_2}}, \mathcal{X}_{\mathcal{H}_2}(v') = \mathcal{X}_{\mathcal{H}_1}(f'^{-1}(v'))$.
\end{proposition}

The proof is shown in Appendix~A. Furthermore, we conclude that
the dual isomorphism $f'$ satisfies $\forall v \in \mathcal{V}_{\mathcal{H}_1}$, $f'(v) = g_2((f(g_1^{-1}(v).\text{source}), f(g_1^{-1}(v).\text{target})))$.
We denote $\mathcal{G}_1 \simeq \mathcal{G}_2 \rightarrow L(\mathcal{G}_1) \simeq L(\mathcal{G}_2)$ for Proposition~\ref{propo:line_iso}.



\begin{figure}[!tb]
    \centering
    \begin{subfigure}{.23\textwidth}
        \centering
        \includegraphics[height=3.5cm]{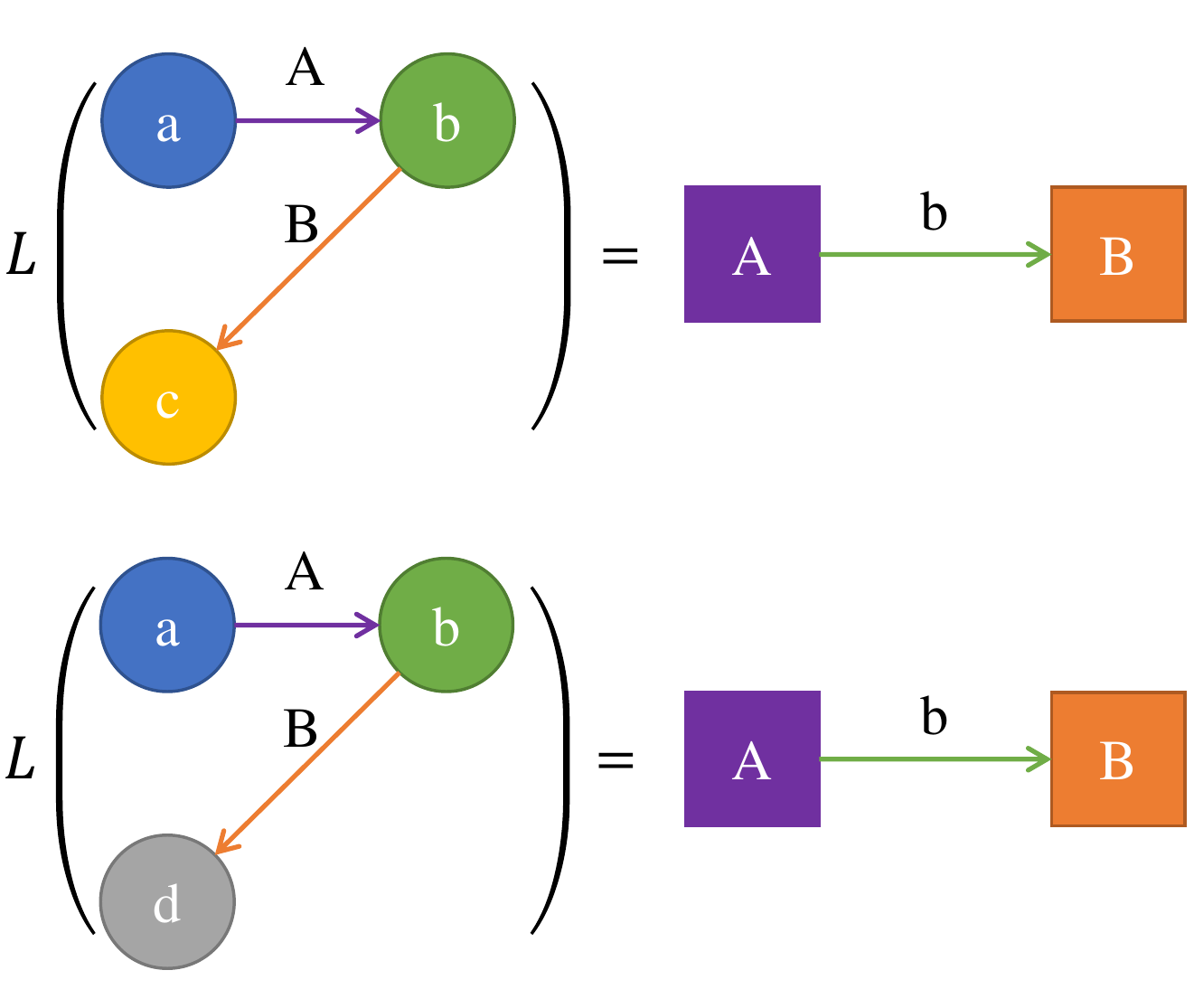}
        \vspace{-0.15in}
        \caption{Non-injective case}
        \label{fig:non4}
    \end{subfigure}
    \begin{subfigure}{.01\textwidth}
    \end{subfigure}
    \begin{subfigure}{.23\textwidth}
        \centering
        \includegraphics[height=3.5cm]{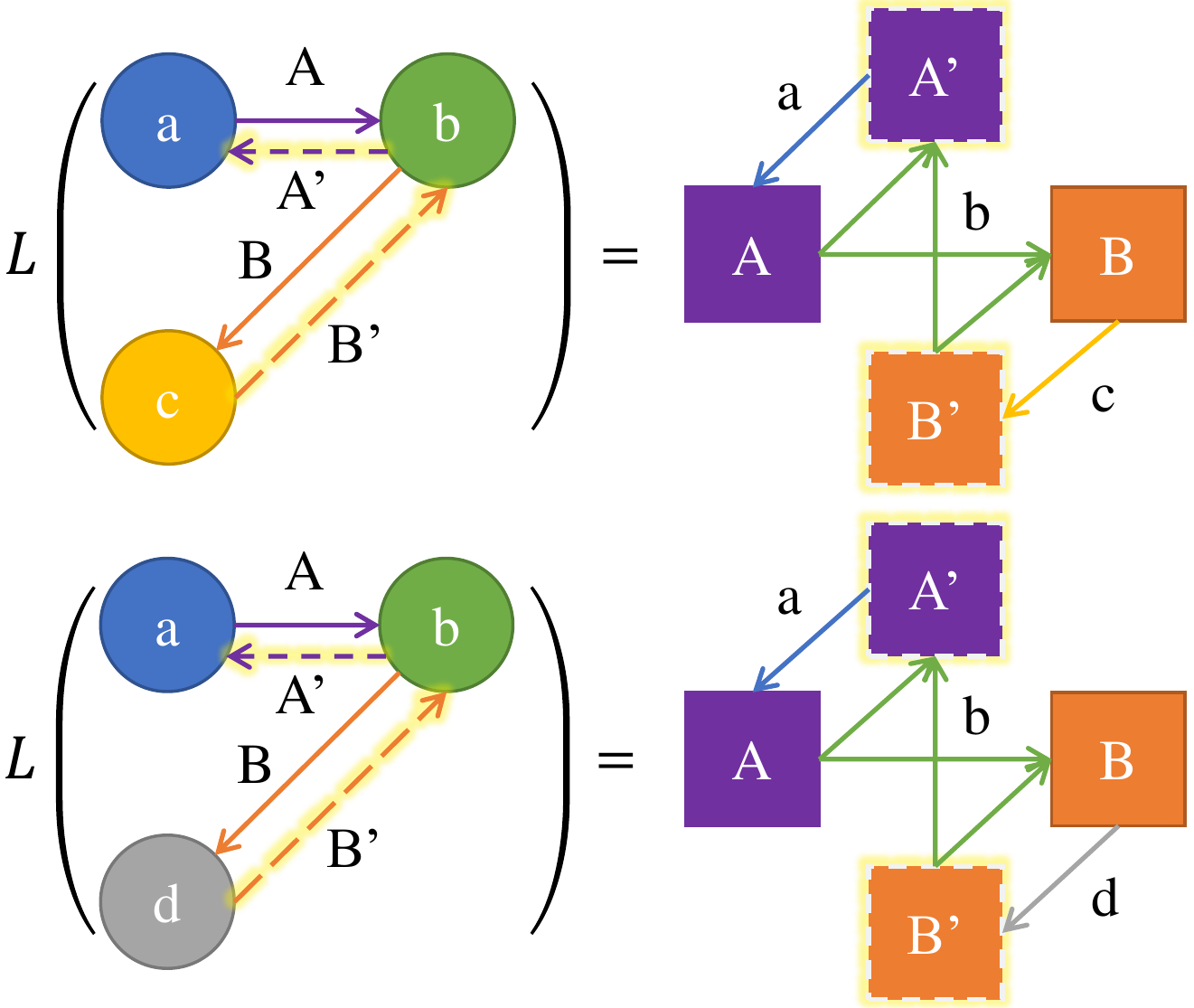}
        \vspace{-0.15in}
        \caption{Adding reversed edges}
        \label{fig:non5}
    \end{subfigure}
    \vspace{-0.1in}
    \caption{Non-isomorphic graphs and their line graphs.}
    \vspace{-0.15in}
    \label{fig:non}
\end{figure}

The relation between the isomorphism $f$ and its dual $f'$ is non-injective: two line graphs in Figure~\ref{fig:non4} are isomorphic but their original graphs are not, which also indicates $f'$ may correspond to multiple different $f$ (even $f$ does not exist).
That is to say, the edge-to-vertex transform $L$ cannot remain \revisexin{all graph adjacency} and guarantee isomorphisms in some situations.


\begin{theorem}[Whitney isomorphism theorem]
\label{theorem:whitney_iso}
For connected simple graphs with more than four vertices, there is a one-to-one correspondence between isomorphisms of the graphs and isomorphisms of their line graphs.
\end{theorem}
Theorem~\ref{theorem:whitney_iso}~\cite{whitney1932congruent} concludes the condition for simple graphs.
Inspired by it,
we add reversed edges associated with \revisexin{special} labels for directed graphs so that graphs can be regarded as undirected (Figure~\ref{fig:non5}).
Theorem~\ref{theorem:ext_whitney_iso} is the extension for directed heterogeneous multigraphs.

\begin{theorem}
\label{theorem:ext_whitney_iso}
For connected directed heterogeneous multigraphs with reversed edges (the \reviseyq{reverse} of one self-loop is itself), there is a one-to-one correspondence between isomorphisms of the graphs and isomorphisms of their line graphs.
\end{theorem}

The detailed proof is listed in Appendix~B. Moreover, we have Corollary~\ref{corollary:subiso_equal} for subgraph isomorphisms and their duals.

\begin{corollary}
{For connect directed heterogeneous multigraphs with reversed edges more than one vertex, there is a one-to-one correspondence between subgraph isomorphisms of the graphs and subgraph isomorphisms of their line graphs.}\label{corollary:subiso_equal}
\end{corollary}

\section{Dual Message Passing Neural Networks}
\revisexin{
The edge-to-vertex transform and the duality property indicate that searching isomorphisms on the original graph is equivalent to searching on its line graph.
Hence, we design the dual message passing to model nodes with original structure and model edges with the line graph structure.
Moreover, we extend the dual message passing to heterogeneous multi-graphs.}

\subsection{Conventional Graph Convolutions}
\citeauthor{kipf2017semi}~(\citeyear{kipf2017semi}) proposed parameterized conventional graph convolutions as the first-order approximation of spectral convolutions $\bm \Theta \star \bm h = \bm U \bm \Theta \bm U^{\top} \bm h$, where $\bm \Theta$ is the filter in the Fourier domain and $\bm h \in \mathbb{R}^{n}$ is the scalar feature vector for $n$ vertices of $\mathcal{G}$.
In practice, $\bm \Theta$ is a diagonal matrix as a function of eigenvalues of the (normalized) graph Laplacian.
Considering the computational cost of eigendecomposition is $\mathcal{O}(n^3)$, it is approximated by shifted Chebyshev polynomials~\cite{hammond2011wavelets}:
\scalebox{0.90}{\parbox{1.111\linewidth}{
\begin{align}
    \bm \Theta &\approx \sum_{k=0}^{K}{T_k(\frac{2}{\lambda_{{\mathcal{G}}{\text{max}}}}\bm \Lambda_{\mathcal{G}} - \bm I_{n}) \bm \theta_k} \nonumber \\
    &\approx T_0(\frac{2}{\lambda_{{\mathcal{G}}{\text{max}}}}\bm \Lambda_{\mathcal{G}} - \bm I_{n}) \bm \theta_0 + T_1(\frac{2}{\lambda_{{\mathcal{G}}{\text{max}}}}\bm \Lambda_{\mathcal{G}} - \bm I_{n}) \bm \theta_1 \nonumber \\
    &=\bm \theta_0 + (\frac{2}{\lambda_{{\mathcal{G}}{\text{max}}}}\bm \Lambda_{\mathcal{G}} - \bm I_{n})\bm \theta_1,
\end{align}
}}
where $\bm \Lambda_{\mathcal{G}}$ is the diagonal matrix of eigenvalues, $\bm I_{n} \in \mathbb{R}^{n \times n}$ is an identity matrix, $\lambda_{{\mathcal{G}}{\text{max}}}$ is the largest eigenvalue so that the input of $T_k(\cdot)$ is located in $[-1, 1]$.
Therefore, the convolution becomes to
\scalebox{0.90}{\parbox{1.111\linewidth}{
\begin{align}
    \bm \Theta \star \bm h &= \bm U \bm \Theta \bm U^T \bm h \nonumber \\
    &\approx \bm U \bm (\bm \theta_0 + (\frac{2}{\lambda_{{\mathcal{G}}{\text{max}}}}\bm \Lambda_{\mathcal{G}} - \bm I_{n \times n})\bm \theta_1) \bm U^T \bm h \nonumber \\
    &= (\bm \theta_0 - \bm \theta_1) \bm h + \frac{2 \bm \theta_1}{\lambda_{{\mathcal{G}}{\text{max}}}} \bm{L}_{\mathcal{G}} \bm h, \label{eq:node_conv}
\end{align}
}}
where $\bm L_{\mathcal{G}}$ is the (normalized) graph Laplacian matrix.
$\lambda_{\mathcal{G}\text{max}}$ is bounded by $\text{max}\{d_u + d_v | (u, v) \in \mathcal{E}_\mathcal{G}\}$ if the Laplacian $\bm{L}_{\mathcal{G}} = \bm{D}_{\mathcal{G}} - \bm{A}_{\mathcal{G}}$ or by $2$ if the Laplacian is noramlzied as $\bm{D}_{\mathcal{G}}^{-\frac{1}{2}} (\bm{D}_{\mathcal{G}} - \bm{A}_{\mathcal{G}}) \bm{D}_{\mathcal{G}}^{-\frac{1}{2}} = \bm{I}_{n} - \bm{D}_{\mathcal{G}}^{-\frac{1}{2}}\bm{A}_{\mathcal{G}}\bm{D}_{\mathcal{G}}^{-\frac{1}{2}}$, where $\bm{A}_{\mathcal{G}}$ is the adjacency matrix and $a_{uv}$ corresponds to the number of edges from vertex $u$ to vertex $v$, $\bm{D}_{\mathcal{G}}$ is the degree diagonal matrix and $d_{v}$ is the (out-)degree of vertex $v$~\cite{zhang2011laplacian}.
Graph convolution networks have shown great success in many fields, including node classification, graph property prediction, graph isomorphism test, and subgraph isomorphism counting.
\citeauthor{xu2019how}~(\citeyear{xu2019how}) and \citeauthor{liu2020neural}~(\citeyear{liu2020neural}) found that the sum aggregation is good at capturing structural information and solving isomorphism problems.
Hence, we consider to use the unnormalized graph Laplacian $\bm{L}_{\mathcal{G}} = \bm{D}_{\mathcal{G}} - \bm{A}_{\mathcal{G}}$ and set $\lambda_{\mathcal{G}\text{max}} = \text{max}\{d_u + d_v | (u, v) \in \mathcal{E}_\mathcal{G}\}$.

\subsection{Dual Message Passing Mechanism}
This convolution can also apply on the line graph $\mathcal{H} = L(\mathcal{G})$, then convolutional operation in $\mathcal{H}$ is
\scalebox{0.90}{\parbox{1.111\linewidth}{
\begin{align}
    \bm \Gamma \star \bm z \approx (\bm \gamma_0 - \bm \gamma_1) \bm z + \frac{2 \bm \gamma_1}{\lambda_{{\mathcal{H}}{\text{max}}}} \bm{L}_{\mathcal{H}} \bm z, \label{eq:edge_conv}
\end{align}
}}
where $\bm \Gamma$ is the filter for $\mathcal{H}$, $\bm z \in \mathbb{R}^{m}$ is the scalar feature vector for $m$ vertices of $\mathcal{H}$, and $\lambda_{\mathcal{H}\text{max}}$ is the largest eigenvalue of the Laplacian $L_{\mathcal{H}}$, which is no greater than $\text{max}\{d_u + d_v | (u, v) \in \mathcal{E}_{H}\}$.
We can use Eq.~(\ref{eq:edge_conv}) to acquire the edge representations of $\mathcal{G}$ because Definition~\ref{def:line} and Corollary~\ref{corollary:subiso_equal} show the line graph $\mathcal{H}$ also preserves the structural information of $\mathcal{G}$ for subgraph isomorphisms.

However, Eq.~(\ref{eq:edge_conv}) results \reviseyq{in} a new problem: the computation cost is linear to $|\mathcal{E}_{\mathcal{H}}| = \frac{1}{2}\sum_{v \in \mathcal{G}}{d_{v}^2} - n = \mathcal{O}(m^2)$ where $m=|\mathcal{E}_\mathcal{G}|=|\mathcal{V}_\mathcal{H}|$.
To tackle this issue, we combine the two convolutions in an asynchronous manner in $\mathcal{O}(m)$.

\begin{proposition}
\label{proposition:incidence_laplacian}
    If $\mathcal{G}$ is a directed graph with $n$ vertices and $m$ edges, then $\bm{A}_\mathcal{G} + \bm{A}^\top_\mathcal{G} = \bm{D}^{+}_\mathcal{G} + \bm{D}^{-}_\mathcal{G} - \bm{B}_\mathcal{G}\bm{B}_\mathcal{G}^\top$, where $\bm{A}_\mathcal{G} \in \mathbb{R}^{n \times n}$ is the adjacency matrix, $\bm{D}^{+}_\mathcal{G}, \bm{D}^{-}_\mathcal{G} \in \mathbb{R}^{n \times n}$ are the out-degree and in-degree diagonal matrices respectively, and $\bm{B}_\mathcal{G} \in \mathbb{R}^{n \times m}$ is the oriented incidence matrix where $b_{ve}=1$ if vertex $v$ is the destination of edge $e$, $b_{ve}=-1$ if $v$ is the source of $e$, $b_{ve}=0$ otherwise. 
    In \reviseyq{particular}, if $\mathcal{G}$ is with reversed edges, then we have $\bm{B}_\mathcal{G}\bm{B}_\mathcal{G}^\top = 2(\bm{D}_\mathcal{G} - \bm{A}_\mathcal{G}) = 2 \bm{L}_\mathcal{G}$, where $\bm{L}_\mathcal{G} \in \mathbb{R}^{n \times n}$ is the Laplacian matrix.
\end{proposition}

\begin{proposition}
\label{proposition:line_adj}
    If $\mathcal{G}$ is a directed graph with $n$ vertices and $m$ edges and $\mathcal{H}$ is the line graph of $\mathcal{G}$, then $\bm{A}_{\mathcal{H}} + \bm{A}_{\mathcal{H}}^\top = \bm{\hat{B}}_{\mathcal{G}}^\top\bm{\hat{B}}_{\mathcal{G}} - 2\bm{I}_{m}$,
    where $\bm{A}_\mathcal{H} \in \mathbb{R}^{m \times m}$ is the adjacency matrix of $\mathcal{H}$, $\bm{I}_{m} \in \mathbb{R}^{m \times m}$ is an identity matrix, and $\bm{\hat{B}}_\mathcal{G} \in \mathbb{R}^{n \times m}$ is the unoriented incidence matrix where $\hat{b}_{ve}=1$ if vertex $v$ is incident to edge $e$, $\hat{b}_{ve}=0$ otherwise. 
    In \reviseyq{particular}, if $\mathcal{G}$ is with reversed edges, then $\mathcal{H}$ is also with reversed edges and $\bm{A}_{\mathcal{H}} = \frac{1}{2}\bm{\hat{B}}_{\mathcal{G}}^\top\bm{\hat{B}}_{\mathcal{G}} - \bm{I}_{m}$.
    Furthermore, we have $\bm{L}_\mathcal{H} = \bm{D}_\mathcal{H} - \bm{A}_\mathcal{H} = \bm{D}_\mathcal{H} + \bm{I}_{m} - \frac{1}{2}\bm{\hat{B}}_{\mathcal{G}}^\top\bm{\hat{B}}_{\mathcal{G}}$, where $\bm{L}_\mathcal{H} \in \mathbb{R}^{m \times m}$ is the Laplacian matrix of $\mathcal{H}$.
\end{proposition}

We use Proposition~\ref{proposition:incidence_laplacian} to inspect the graph convolutions.
The second term of \reviseyq{Eq.~(\ref{eq:node_conv})} can be written as $\frac{\bm \theta_1}{\lambda_{{\mathcal{G}}{\text{max}}}} \bm{B}_{\mathcal{G}}\bm{B}_{\mathcal{G}}^\top \bm h$, and $\bm{B}_{\mathcal{G}}^\top \bm h \in \mathbb{R}^{m}$ corresponds to the computation $\{x_{v} - x_{u}|(u,v) \in \mathcal{E}_{\mathcal{G}}\}$ in the edge space.
We can design a better filter to replace this subtraction operation so that $\bm{\Theta} \star \bm h \approx (\bm \theta_0 - \bm \theta_1) \bm h + \frac{\bm \theta_1}{\lambda_{{\mathcal{G}}{\text{max}}}} \bm{B}_{\mathcal{G}} \bm z$, where $z$ is the result of some specific computation in the edge space, which is straightforward to involve Eq.~(\ref{eq:edge_conv}).
We are able to generalize Eq.~(\ref{eq:edge_conv}) by the same idea, but it does not help to reduce the complexity. The second term of Eq.~(\ref{eq:edge_conv}) is equivalent to $\frac{2 \bm \gamma_1}{\lambda_{{\mathcal{H}}{\text{max}}}}(\bm{D}_{\mathcal{H}} + \bm{I}_{m})\bm{z} - \frac{\bm \gamma_1}{\lambda_{{\mathcal{H}}{\text{max}}}}\bm{\hat{B}}_{\mathcal{G}}^\top\bm{\hat{B}}_{\mathcal{G}}\bm{z}$ obtained from Proposition~\ref{proposition:line_adj}.
\reviseyq{Moreover}, $\bm{\hat{B}}_{\mathcal{G}}\bm{z} \in \mathbb{R}^{n}$ corresponds to the computation $\{\sum_{(u,v) \in {\mathcal{E}_{\mathcal{G}}}} \bm{z}_{uv} + \sum_{(v,u) \in {\mathcal{E}_{\mathcal{G}}}} \bm{z}_{vu}| v \in \mathcal{V}_{\mathcal{G}}\}$.
We can also enhance this computation by introducing $\bm{h}$, e.g., $\bm{\Gamma} \star \bm z \approx (\bm \gamma_0 - \bm \gamma_1) \bm z + \frac{2 \bm \gamma_1}{\lambda_{\mathcal{H}\text{max}}}(\bm{D}_{\mathcal{H}} + \bm{I}_{m}) \bm z - \frac{\bm \gamma_1}{\lambda_{{\mathcal{H}}{\text{max}}}} \bm{\hat{B}}_{\mathcal{G}}^\top \bm h$.
We can get the degree matrix $\bm{D}_\mathcal{H}$ without constructing the line graph $\mathcal{H}$ because it depends on the vertex degrees of $\mathcal{G}$: $\{d_{g(e)}^- = d_u^-, d_{g(e)}^+ = d_v^+| e = (u, v) \in \mathcal{E}_\mathcal{G}\}$.
\revisexin{We manually set $\lambda_{\mathcal{H}_{\text{max}}} = \text{max}\{d_u + d_v | (u, v) \in \mathcal{E}_\mathcal{H}\} = \text{max}\{d_u^- + d_v^+ | (u, v) \in \mathcal{E}_\mathcal{G}\}$.}

Finally, the asynchronous updates are defined as follows:
\scalebox{0.90}{\parbox{1.111\linewidth}{
\begin{align}
    \bm h^{(k)} \leftarrow (\bm \theta^{(k)}_0 - \bm \theta^{(k)}_1) \bm h^{(k-1)} &+ \frac{\bm \theta^{(k)}_1}{\lambda_{{\mathcal{G}}{\text{max}}}} \bm{B}_{\mathcal{G}} \bm z^{(k-1)}, \label{eq:homo_node_update} \\
    \bm z^{(k)} \leftarrow (\bm \gamma^{(k)}_0 - \bm \gamma^{(k)}_1) \bm z^{(k-1)} 
    &+ \frac{2 \bm \gamma^{(k)}_1}{\lambda_{\mathcal{H}\text{max}}}(\bm{D}_{\mathcal{H}} + \bm{I}_{m}) \bm z^{(k-1)} \nonumber\\
    &- \frac{\bm \gamma^{(k)}_1}{\lambda_{{\mathcal{H}}{\text{max}}}} \bm{\hat{B}}_{\mathcal{G}}^\top \bm h^{(k-1)}, \label{eq:homo_edge_update}
\end{align}
}}
where $\bm{\theta}_{:}^{(k)}$ and $\bm{\gamma}_{:}^{(k)}$ indicate the parameters at the $k$-th update and $\bm{h}^{(k)}$ and $\bm{z}^{(k)}$ are the updated results.
The computation of $\bm{B}_{\mathcal{G}} \bm{z}^{(k)}$ and the computation of $\bm{\hat{B}}_{\mathcal{G}}^\top \bm{h}^{(k)}$ are linear to the number of edges $m$ with the help of sparse representations for $\bm{B}_{\mathcal{G}}$ and $\bm{\hat{B}}_{\mathcal{G}}$.

\subsection{Heterogeneous Multi-graph Extensions}
Different relational message passing variants have been proposed to model heterogeneous graphs.
Nevertheless, our dual message passing is natural to handle complex edge types and even edge features.
Each edge not only carries the edge-level property, but \reviseyq{also} stores the local structural information in the corresponding line graph.
However, Eq.~(\ref{eq:homo_edge_update}) does not reflect the edge direction since $\bm{\hat{B}}_{\mathcal{G}}$ regards the source and the target of one edge as the same.
Therefore, we extend Eqs.~(\ref{eq:homo_node_update}-\ref{eq:homo_edge_update}) and propose dual message passing neural networks (DMPNNs) to support the mixture of various properties:
\scalebox{0.90}{\parbox{1.111\linewidth}{
\begin{align}
    \bm H^{(k)} &= \bm H^{(k-1)} \bm{W}^{(k)}_{\theta_0} - (\bm{\hat{B}}_{\mathcal{G}} - \bm{B}_{\mathcal{G}}) \bm Z^{(k-1)}\bm{W}^{(k)}_{\theta_1^-} \nonumber \\
    &\qquad\qquad\qquad\quad + (\bm{\hat{B}}_{\mathcal{G}} + \bm{B}_{\mathcal{G}}) \bm Z^{(k-1)}\bm{W}^{(k)}_{\theta_1^+}, \label{eq:hete_node_update} \\
    \bm Z^{(k)} &= \bm Z^{(k-1)} \bm{W}^{(k)}_{\gamma_0} + 2(\bm{D}_{\mathcal{H}} + \bm{I}_{m}) \bm Z^{(k-1)} (\bm{W}^{(k)}_{\gamma_1^-} - \bm{W}^{(k)}_{\gamma_1^+}) \nonumber \\
    &\qquad\qquad\qquad\quad - (\bm{\hat{B}}_{\mathcal{G}} - \bm{B}_{\mathcal{G}})^\top \bm H^{(k-1)} \bm{W}^{(k)}_{\gamma_1^-} \nonumber \\
    &\qquad\qquad\qquad\quad + (\bm{\hat{B}}_{\mathcal{G}} + \bm{B}_{\mathcal{G}})^\top \bm H^{(k-1)} \bm{W}^{(k)}_{\gamma_1^+}, \label{eq:hete_edge_update}
\end{align}
}}
where 
$\bm{H}^{(k)} \in \mathbb{R}^{n \times l^{(k)}}$ and $\bm{Z}^{(k)} \in \mathbb{R}^{m \times l^{(k)}}$ are $l^{(k)}$-dim hidden states of nodes and edges in the $k$-th DMPNN layer.
$\bm{H}^{(0)}$ and $\bm{Z}^{(0)}$ are initialized with features, labels, and other properties,
$\bm{\hat{B}}_{\mathcal{G}} - \bm{B}_{\mathcal{G}}$ eliminates out-edges,
$\bm{\hat{B}}_{\mathcal{G}} + \bm{B}_{\mathcal{G}}$ filters out in-edges,
and $\bm{W}_{\theta_:}^{(k)}$ and $\bm{W}_{\gamma_:}^{(k)} \in \mathbb{R}^{l^{(k-1)} \times l^{(k)}}$ are trainable parameters \revisexin{that are initialized bounded by $\frac{\sqrt{\frac{6}{l^{(k-1)} + l^{(k)}}}}{{\lambda_{{\mathcal{G}}{\text{max}}}}}$ and $\frac{\sqrt{\frac{6}{l^{(k-1)} + l^{(k)}}}}{{\lambda_{{\mathcal{H}}{\text{max}}}}}$, respectively}.
For the detailed explanations and reparameterization tricks, see Appendix~C.
After $K$ updates, we finally get $l^{(K)}$-dim node and edge representations $\bm{H}^{(K)}$ and  $\bm{Z}^{(K)}$ in the aligned space.

\section{Experiments}
We evaluate DMPNNs on the challenging subgraph isomorphism counting and matching tasks.
\revisexin{Besides, we also learn embeddings and classify nodes without any label or attribute on heterogeneous graphs to verify the generalization and the necessity of explicit edge modeling.}
Training and testing of DMPNNs and baselines were \reviseyq{conducted} on single NVIDIA V100 GPU under 
PyTorch~\cite{paszke2019pytorch} and DGL~\cite{wang2019dgl} frameworks.


\subsection{Subgraph Isomorphism Counting and Matching}
DMPNNs are designed based on the duality of isomorphisms so that evaluation on isomorphism related tasks is the most straightforward.
Given a pair of pattern $\mathcal{P}$ and graph $\mathcal{G}$, subgraph isomorphism counting aims to count all different subgraph isomorphisms in $\mathcal{G}$, and matching aims to seek out which nodes and edges belong to those isomorphic subgraphs.
We report the root mean square error (RMSE) and the mean absolute error (MAE) between global counting predictions and the ground truth, and evaluate graph edit distance (GED) between predicted subgraphs and all isomorphic subgraphs.
\revisexin{However, computing GED is NP-hard, so we consider the lower-bound of GED in contiguous space.}
We \reviseyq{use} DMPNN and baselines to predict the possible frequency of each node or edge appearing in isomorphic subgraphs.
For example, models are expected to return $[2, 2, 2]$ for nodes and $[2,2,2]$ for edges given the pair in Figure~\ref{fig:preliminary_iso}, and return $[1, 1, 1, 0]$ for nodes and $[1, 1, 1, 0, 0]$ for edges given  Figure~\ref{fig:preliminary_subiso}.
MAE between node predictions and node frequencies or the MAE between edge predictions and edge frequencies is regarded as the lower-bound of GED.
We run experiments on three different seeds and report the best.

\subsubsection{Models}
We compare with three sequence models and three graph models, including CNN~\cite{kim2014convolutional}, LSTM~\cite{hochreiter1997long}, TXL~\cite{transformerxl2019dai}, RGCN~\cite{schlichtkrull2018modeling}, RGIN~\cite{liu2020neural}, and CompGCN~\cite{vashishth2020composition}.
Sequence models embed edges, and we calculate the MAE over edges as the GED.
On the contrary, graph models embed nodes so that we consider the MAE over nodes.
We jointly train counting and matching prediction modules of DMPNN and other graph baselines:
\scalebox{0.87}{\parbox{1.149\linewidth}{
\begin{align}
    \mathcal{J} = \frac{1}{|\mathcal{D}|} \sum_{(\mathcal{P}, \mathcal{G}) \in \mathcal{D}}{\Bigl(\left(c_{\mathcal{P}, \mathcal{G}} - p_{\mathcal{P}, \mathcal{G}}\right)^2 + \sum_{v \in \mathcal{V}_{\mathcal{G}}}{\left(w_{\mathcal{P}, v} - p_{\mathcal{P}, v}\right)^2}\Bigr)}, \label{eq:obj_sic}
\end{align}
}}
where $\mathcal{D}$ is the dataset containing pattern-graph pairs, $c_{\mathcal{P}, \mathcal{G}}$ indicates the ground truth of number of subgraph isomorphisms between pattern $\mathcal{P}$ and graph $\mathcal{G}$, $c_{\mathcal{P}, v}$ indicates the frequency of vertex $v$ appearing in isomorphisms, $p_{\mathcal{P}, \mathcal{G}} $ \reviseyq{and} $ p_{\mathcal{P}, v}$ are the corresponding predictions.
\reviseyq{For sequence models, we} jointly minimize the MSE of counting predictions and the MSE of edge predictions.
We follow the same setting of \citeauthor{liu2020neural}~(\citeyear{liu2020neural}) to combine multi-hot encoding and message passing to embed graphs and use pooling operations to make predictions:
\scalebox{0.839}{\parbox{1.189\linewidth}{
\begin{align}
    \bm{h}_{v}^{(0)} &= \text{Concat}(\text{MultiHot}(v), \text{MultiHot}(\mathcal{X}(v))) \bm{W}_{\text{vertex}}, \nonumber \\
    \bm{z}_{e}^{(0)} &= \text{MultiHot}(\mathcal{Y}(e)) \bm{W}_{\text{edge}}, \nonumber \\
    \bm{H}^{(K)}, \bm{Z}^{(K)} &= \text{DMPNN}^{(K)}(\cdots(\text{DMPNN}^{(1)}(\bm{H}^{(0)}, \bm{Z}^{(0)}))), \nonumber \\
    \bm{p} &= \sum_{v \in {\mathcal{P}}}{\bm{h}^{(K)}_{\mathcal{P}v}}, \quad \bm{g} = \sum_{v \in {\mathcal{G}}}{\bm{h}^{(K)}_{\mathcal{G}v}}, \nonumber
\end{align}
}}
\scalebox{0.87}{\parbox{1.149\linewidth}{
\begin{align}
    p_{\mathcal{P}, v} &= \text{FC}_{\text{matching}}(\text{Concat}(\bm{h}^{(K)}_{\mathcal{G}v}, \bm p, \bm{h}^{(K)}_{\mathcal{G}v}-\bm p, \bm{h}^{(K)}_{\mathcal{G}v}\odot\bm p)), \nonumber\\
    p_{\mathcal{P}, \mathcal{G}} &= \text{FC}_{\text{counting}}(\text{Concat}(\bm g, \bm p, \bm g-\bm p, \bm g\odot\bm p)), \nonumber
\end{align}
}}
where $\bm{W}_{\text{vertex}}$ and $\bm{W}_{\text{edge}}$ are trainable matrices to align id and label representations to the same dimension.
We also consider the more powerful Deep-LRP~\cite{chen2020can} and add local relational pooling behind dual message passing for node representation learning, denoted as DMPNN-LRP.
For \revisexin{a} fair comparison, we use 3-layer networks and set the embedding dimensions, hidden sizes, and numbers of filters as 64 for all models.
We follow the original setting of Deep-LRP to use 3-truncated BFS. 
Considering the quadratic computation complexity of TXL, we set the segment size and memory size as 128.
All models are trained using AdamW~\cite{loshchilov2019} with a learning rate 1e-3 and a decay 1e-5.

\subsubsection{Datasets}
Table~\ref{table:stat_sic} shows the statistics of two synthetic homogeneous datasets with 3-stars, triangles, tailed triangles, and chordal cycles as patterns ~\cite{chen2020can}, one synthetic heterogeneous dataset with 75 random patterns,\footnote{This \textit{Complex} dataset corresponds to the \textit{Small} dataset in the original paper. But we found some ground truth counts are not correct because VF2 does not check self-loops. We removed all self-loops from patterns and graphs and got the correct ground truth.} and one mutagenic compound dataset \textit{MUTAG} with 24 patterns~\cite{liu2020neural}.
In traditional algorithms, adding reversed edges increases the search space dramatically, but it does not take too much extra time on neural methods.
Thus, we also conduct experiments on patterns and graphs with reversed edges associated with specific edge labels
, which doubles the number of edges and the number of edge labels.

\begin{table}[t]
    \footnotesize
    \centering
    \setlength\tabcolsep{1.2pt}
    \resizebox{\linewidth}{!}{%
    \begin{tabular}{l|ccc|ccc|ccc|ccc}
    \toprule
        & \multicolumn{3}{c|}{Erd\H{o}s-Renyi} & \multicolumn{3}{c|}{Regular} & \multicolumn{3}{c|}{Complex} & \multicolumn{3}{c}{MUTAG} \\
        \toprule
        \#train & \multicolumn{3}{c|}{6,000} & \multicolumn{3}{c|}{6,000} & \multicolumn{3}{c|}{358,512} & \multicolumn{3}{c}{1,488} \\
        \#valid & \multicolumn{3}{c|}{4,000} & \multicolumn{3}{c|}{4,000} & \multicolumn{3}{c|}{44,814} & \multicolumn{3}{c}{1,512} \\
        \#test & \multicolumn{3}{c|}{10,000} & \multicolumn{3}{c|}{10,000} & \multicolumn{3}{c|}{44,814} & \multicolumn{3}{c}{1,512} \\
        \midrule
        & Max & \multicolumn{2}{c|}{Avg} & Max & \multicolumn{2}{c|}{Avg} & Max & \multicolumn{2}{c|}{Avg} & Max & \multicolumn{2}{c}{Avg} \\
        $|\mathcal{V}_{\mathcal{P}}|$ & 4 & \multicolumn{2}{c|}{3.8$\pm$0.4} & 4 & \multicolumn{2}{c|}{3.8$\pm$0.4} & 8 & \multicolumn{2}{c|}{5.2$\pm$2.1} & 4 & \multicolumn{2}{c}{3.5$\pm$0.5} \\
        $|\mathcal{E}_{\mathcal{P}}|$ & 10 & \multicolumn{2}{c|}{7.5$\pm$1.7} & 10 & \multicolumn{2}{c|}{7.5$\pm$1.7} & 8 & \multicolumn{2}{c|}{5.9$\pm$2.0} & 3 & \multicolumn{2}{c}{2.5$\pm$0.5} \\
        $|\mathcal{X}_{\mathcal{P}}|$ & 1 & \multicolumn{2}{c|}{1$\pm$0} & 1 & \multicolumn{2}{c|}{1$\pm$0} & 8 & \multicolumn{2}{c|}{3.4$\pm$1.9} & 2 & \multicolumn{2}{c}{1.5$\pm$0.5} \\
        $|\mathcal{Y}_{\mathcal{P}}|$ & 1 & \multicolumn{2}{c|}{1$\pm$0} & 1 & \multicolumn{2}{c|}{1$\pm$0} & 8 & \multicolumn{2}{c|}{3.8$\pm$2.0} & 2 & \multicolumn{2}{c}{1.5$\pm$0.5} \\
        $|\mathcal{V}_{\mathcal{G}}|$ & 10 & \multicolumn{2}{c|}{10$\pm$0} & 30 & \multicolumn{2}{c|}{18.8$\pm$7.4} & 64 & \multicolumn{2}{c|}{32.6$\pm$21.2} & 28 & \multicolumn{2}{c}{17.9$\pm$4.6} \\
        $|\mathcal{E}_{\mathcal{G}}|$ & 48 & \multicolumn{2}{c|}{27.0$\pm$6.1} & 90 & \multicolumn{2}{c|}{62.7$\pm$17.9} & 256 & \multicolumn{2}{c|}{73.6$\pm$66.8} & 66 & \multicolumn{2}{c}{39.6$\pm$11.4} \\
        $|\mathcal{X}_{\mathcal{G}}|$ & 1 & \multicolumn{2}{c|}{1$\pm$0} & 1 & \multicolumn{2}{c|}{1$\pm$0} & 16 & \multicolumn{2}{c|}{9.0$\pm$4.8} & 7 & \multicolumn{2}{c}{3.3$\pm$0.8} \\
        $|\mathcal{Y}_{\mathcal{G}}|$ & 1 & \multicolumn{2}{c|}{1$\pm$0} & 1 & \multicolumn{2}{c|}{1$\pm$0} & 16 & \multicolumn{2}{c|}{9.4$\pm$4.7} & 4 & \multicolumn{2}{c}{3.0$\pm$0.1} \\
        \bottomrule
    \end{tabular}
    }
    \vspace{-0.1in}
    \caption{Statistics of datasets on subgraph isomorphism experiments. $\mathcal{P}$ and $\mathcal{G}$ corresponds to patterns and graphs.}
    \label{table:stat_sic}
    \vspace{-0.3in}
\end{table}

\begin{table*}[t]
    \footnotesize
    \vspace{-0.1in}
    \centering
    \setlength\tabcolsep{4pt}
    \begin{tabular}{l|ccc|ccc|ccc|ccc}
    \toprule
        \multicolumn{1}{c|}{\multirow{3}{*}{Models}} & \multicolumn{6}{c|}{Homogeneous}  & \multicolumn{6}{c}{Heterogeneous} \\
        \cline{2-13}
        & \multicolumn{3}{c|}{Erd\H{o}s-Renyi} & \multicolumn{3}{c|}{Regular} & \multicolumn{3}{c|}{Complex} & \multicolumn{3}{c}{MUTAG} \\
        & RMSE & MAE & GED & RMSE & MAE & GED & RMSE & MAE & GED & RMSE & MAE & GED \\
        \midrule
        \multirow{1}{*}{Zero} & 92.532 & 51.655 & 201.852 & 198.218 & 121.647 & 478.990 & 68.460 & 14.827 & 86.661 & 16.336 & 6.509 & 15.462 \\
        \multirow{1}{*}{Avg}  & 121.388 & 131.007  & 237.349 & 156.515 & 127.211 & 576.476 & 66.836 & 23.882 & 156.095 & 14.998 & 10.036 & 27.958 \\
        \hline
        \multirow{1}{*}{CNN}
        & 20.386 & 13.316 & NA & 37.192 & 27.268 & NA & 41.711 & 7.898 & NA & 1.789 & 0.734 & NA  \\
        \multirow{1}{*}{LSTM}
        & 14.561 & 9.949 & 160.951 & 14.169 & 10.064 & 234.351 & 30.496 & 6.839 & 88.739 & 1.285 & 0.520 & 3.873 \\
        \multirow{1}{*}{TXL}
        & 10.861 & 7.105 & 116.810 & 15.263 & 10.721 & 208.798 & 43.055 & 9.576 & 98.124 & 1.895 & 0.830 & 4.618 \\
        \multirow{1}{*}{RGCN}
        & 9.386 & 5.829 & 28.963 & 14.789 & 9.772 & 70.746 & 28.601 & 9.386 & 64.122 & 0.777 & 0.334 & 1.441 \\
        \multirow{1}{*}{RGIN}
        & 6.063 & 3.712 & \bf 22.155 & 13.554 & 8.580 & 56.353 & 20.893 & 4.411 & 56.263 & 0.273 & 0.082 & 0.329 \\
        \multirow{1}{*}{CompGCN}
        & 6.706 & 4.274 & 25.548 & 14.174 & 9.685 & 64.677 & 22.287 & 5.127 & 57.082 & 0.300 & 0.085 & 0.278 \\
        \multirow{1}{*}{\bf DMPNN}
        & \bf 5.062 & \bf 3.054 & 23.411 & \bf 11.980 & \bf 7.832 & \bf 56.222 & \bf 17.842 & \bf 3.592 & \bf 38.322 & \bf 0.226 & \bf 0.079 & \bf 0.244 \\
        \hline
        \multirow{1}{*}{Deep-LRP}
        & 0.794 & 0.436 & 2.571 & 1.373 & 0.788 & 5.432 & 27.490 & 5.850 & 56.772 & 0.260 & 0.094 & 0.437 \\
        \multirow{1}{*}{\bf DMPNN-LRP}
        & \bf 0.475 & \bf 0.287 & \bf 1.538 & \bf 0.617 & \bf 0.422 & \bf 2.745 & \bf 17.391 & \bf 3.431 & \bf 35.795  & \bf 0.173 & \bf 0.053 & \bf 0.190 \\
        \bottomrule
    \end{tabular}
    \vspace{-0.1in}
    \caption{Performance on subgraph isomorphism counting and matching.}
    \label{table:subisocnt}
    \vspace{-0.15in}
\end{table*}

\begin{table}[t]
    \footnotesize
    \centering
    \setlength\tabcolsep{3pt}
    \resizebox{\linewidth}{!}{%
    \begin{tabular}{rl|ccc|ccc}
    \toprule
        \multicolumn{2}{c|}{\multirow{2}{*}{Models}} & \multicolumn{3}{c|}{Complex} & \multicolumn{3}{c}{MUTAG} \\
        & & RMSE & MAE & GED & RMSE & MAE & GED \\
        \midrule
        \parbox[t]{50pt}{\multirow{2}{*}{CNN}}
        & w/o rev & \bf 41.711 & \bf 7.898 & NA & \bf 1.789 & \bf 0.734 & NA  \\
        & w/ rev  & 47.467 & 10.128 & NA & 2.073 & 0.865 & NA \\
        \hline
        \parbox[t]{50pt}{\multirow{2}{*}{LSTM}}
        & w/o rev & \bf 30.496 & \bf 6.839 & \bf 88.739 & \bf 1.285 & \bf 0.520 & \bf 3.873 \\
        & w/ rev  & 32.178 & 7.575 & 90.718 & 1.776 & 0.835 & 5.744 \\
        \hline
        \parbox[t]{50pt}{\multirow{2}{*}{TXL}}
        & w/o rev & 43.055 & 9.576 & 98.124 & \bf 1.895 & \bf 0.830 & \bf 4.618 \\
        & w/ rev  & \bf 37.251 & \bf 9.156 & \bf 95.887 & 2.701 & 1.175 & 6.436 \\
        \hline
        \parbox[t]{50pt}{\multirow{2}{*}{RGCN}}
        & w/o rev & 28.601 & 9.386 & 64.122 & 0.777 & 0.334 & \bf 1.441 \\
        & w/ rev  & \bf 26.359 & \bf 7.131 & \bf 49.495 & \bf 0.511 & \bf 0.200 & 1.628 \\
        \hline
        \parbox[t]{50pt}{\multirow{2}{*}{RGIN}}
        & w/o rev & 20.893 & 4.411 & 56.263 & 0.273 & \bf 0.082 & \bf 0.329 \\
        & w/ rev  & \bf 20.132 & \bf 4.126 & \bf 39.726 & \bf 0.247 & 0.091 & 0.410 \\
        \hline
        \parbox[t]{50pt}{\multirow{2}{*}{CompGCN}}
        & w/o rev & 22.287 & 5.127 & 57.082 & 0.300 & 0.085 & 0.278 \\
        & w/ rev  & \bf 19.072 & \bf 4.607 & \bf 40.029 & \bf 0.268 & \bf 0.072 & \bf 0.266 \\ 
        \hline
        \parbox[t]{50pt}{\multirow{2}{*}{\bf DMPNN}}
        & w/o rev & 18.974 & 3.922 & 56.933 & 0.232 & 0.088 & 0.320 \\
        & w/ rev  & \bf 17.842 & \bf 3.592 & \bf 38.322 & \bf 0.226 & \bf 0.079 & \bf 0.244 \\
        \hline
        \hline
        \parbox[t]{50pt}{\multirow{2}{*}{Deep-LRP}}
        & w/o rev & 27.490 & 5.850 & \bf 56.772 & \bf 0.260 & \bf 0.094 & \bf 0.437 \\
        & w/ rev  & \bf 26.297 & \bf 5.725 & 61.696 & 0.290 & 0.108 & 0.466 \\
        \hline
        \parbox[t]{50pt}{\multirow{2}{*}{\bf DMPNN-LRP}}
        & w/o rev & 20.425 & 4.173 & 42.200 & 0.196 & 0.062 & 0.210 \\
        & w/ rev  & \bf 17.391 & \bf 3.431 & \bf 35.795 & \bf 0.173 & \bf 0.053 & \bf 0.190 \\
        \bottomrule
    \end{tabular}
    }
    \vspace{-0.1in}
    \caption{Performance comparison after introducing reversed edges on heterogeneous data.}
    \label{table:subisocnt_rev}
    \vspace{-0.2in}
\end{table}

\begin{table}[t]
    \footnotesize
    \centering
    \setlength\tabcolsep{3pt}
    \resizebox{\linewidth}{!}{%
    \begin{tabular}{rl|rr|rr|rr}
    \toprule
        \multicolumn{2}{c|}{\multirow{2}{*}{Models}} & \multicolumn{2}{c|}{MUTAG} & \multicolumn{2}{c|}{Regular} & \multicolumn{2}{c}{Complex} \\
        & & RMSE & MAE & RMSE & MAE & RMSE & MAE \\
        \midrule
        \parbox[t]{50pt}{\multirow{2}{*}{LSTM}}
        & MTL &  1.285 &  0.520 &  14.169 &  10.064 &  30.496 &  6.839 \\
        & STL & -0.003 & \underline{+0.030} & \underline{+0.159} & -0.029 & -1.355 & -0.096 \\
        \hline
        \parbox[t]{50pt}{\multirow{2}{*}{TXL}}
        & MTL &  1.895 &  0.830 &  14.306 &  10.143 &  37.251 &  9.156 \\
        & STL & -0.128 & -0.041 & \underline{+1.487} & \underline{+1.211} & -5.671 & -2.067 \\
        \hline
        \parbox[t]{50pt}{\multirow{2}{*}{RGCN}}
        & MTL &  0.511 &  0.200 &  14.652 &  9.911 &  26.359 &  7.131 \\
        & STL & \underline{+0.202} & \underline{+0.090} & \underline{+0.348} & -0.269 & \underline{+1.686} & \underline{+0.460} \\
        \hline
        \parbox[t]{50pt}{\multirow{2}{*}{RGIN}}
        & MTL &  0.247 &  0.091 &  13.128 &  8.412 &  20.132 &  4.126 \\
        & STL & \underline{+0.053} & \underline{+0.004} & \underline{+1.119} & \underline{+1.019} & \underline{+1.804} & \underline{+0.068} \\
        \hline
        \parbox[t]{50pt}{\multirow{2}{*}{CompGCN}}
        & MTL &  0.268 &  0.072 &  14.174 &  9.685 &  19.072 &  4.607 \\
        & STL & \underline{+0.088} & \underline{+0.086} & \underline{+0.252} & \underline{+0.738} & \underline{+3.625} & \underline{+0.260} \\
        \hline
        \parbox[t]{50pt}{\multirow{2}{*}{\bf DMPNN}}
        & MTL &  0.226 &  0.079 &  11.980 &  7.832 &  17.842 &  3.592 \\
        & STL & \underline{+0.011} & \underline{+0.001} & \underline{+0.318} & \underline{+0.097} & \underline{+3.604} & \underline{+0.865} \\
        \hline
        \hline
        \parbox[t]{50pt}{\multirow{2}{*}{Deep-LRP}}
        & MTL &  0.260 &  0.094 &  1.275 &  0.731 &  26.297 &  5.725 \\
        & STL & \underline{+0.099} & \underline{+0.044} & \underline{+0.036} & \underline{+0.035} & \underline{+3.753} & \underline{+0.886} \\
        \hline
        \parbox[t]{50pt}{\multirow{2}{*}{\bf DMPNN-LRP}}
        & MTL &  0.173 &  0.053 &  0.617 &  0.422 & 17.391 & 3.431 \\
        & STL & \underline{+0.040} & \underline{+0.020} & \underline{+0.513} & \underline{+0.252} & \underline{+4.263} & \underline{+0.928} \\
        \bottomrule
    \end{tabular}
    }
    \vspace{-0.1in}
    \caption{Performance comparison in multi-task training (MTL) and single-task training (STL) on subgraph isomorphism counting. We report best results of whether adding reversed edges or not, and error increases are underlined.}
    \label{table:subisocnt_stl}
    \vspace{-0.2in}
\end{table}

\subsubsection{Results}
Counting and matching results are reported in Table~\ref{table:subisocnt}.
\revisexin{
We find graph models perform better than sequence models, and DMPNN almost surpasses all message passing based networks in counting and matching.
RGIN extends RGCN with the sum aggregator followed by an MLP to makes full use of the neighborhood information, and it improves the original RGCN significantly.
CompGCN is designed to leverage vertex-edge composition operations to predict the potential links, which is contrary to the goal of accurate matching.
On the contrary, DMPNN learns both node embeddings and edge embeddings in aligned space \reviseyq{but from different but dual structures}.
We also observe local relational pooling can significantly decrease errors on homogeneous data by explicitly permuting neighbor subsets.
But Deep-LRP is designed for patterns within three nodes and simple graphs so that it cannot handle multi-edges in nature, let along complex structures in randomly generated data and real-life data.
One advantage of DMPNN is to model heterogeneous nodes and edges in the same space. \reviseyq{We} can see the success of DMPNN-LRP in three \reviseyq{datasets} with the maximum pattern size 4.
But it struggles on the \textit{Complex} \reviseyq{dataset} where patterns contain at most 8 nodes.
}

\revisexin{
We also evaluate baselines with additional reversed edges on \textit{Complex} and \textit{MUTAG} \reviseyq{datasets}.
From results in Table~\ref{table:subisocnt_rev}, we see graph convolutions consistently reduce errors with reversed edges, but sequence models usually become worse.
LRP is designed for simple graphs so that it cannot handle heterogeneous edges in nature, but DMPNN makes it generalized. 
This observation also indicates that one of the challenges on neural subgraph isomorphism counting and matching is the complex graph local structure instead of the number of edges in graphs; otherwise, revised edges were toxic.
}
\revisexin{We compare the efficiency in Appendix~D.}

\revisexin{In the joint learning},
we hope models can learn the mutual supervision that node weights determine the global count and the global count is the upper bound of node weights. 
We also conduct experiments on single task learning to examine whether models can benefit from this mutual supervision.
As shown in Table~\ref{table:subisocnt_stl}, graph models \revisexin{consistently} achieve further performance gains from multi-task learning, while sequence models cannot.
Moreover, improvement is more notable if the dataset is more complicated, e.g., patterns with more edges \revisexin{and graphs with non-trivial structures}.

\subsection{Unattributed Unsupervised Node Classification}
Unattributed unsupervised node classification focuses on local structures instead of node features and attributes.
Node embeddings are learned with the link prediction loss, then linear support vector machines are trained based on 80\% of labeled node embeddings \reviseyq{to} predict the remaining 20\%. We report the average Macro-F1 and Micro-F1 on five runs.

\subsubsection{Models}
We follow the setting of RGCN and CompGCN:
graph neural networks first learn the node representations, and then DistMult models~\cite{embedding2015yang} take pairs of node hidden representations to produce a score for a triplet $\langle u, y, v\rangle$, where $u,y,v$ are the source, the edge type, and the target, respectively.
Eq.~(\ref{eq:obj_unc}) is the objective function, where $\mathcal{D}=\{\langle u,y,v \rangle| (u,v) \in \mathcal{E}_{\mathcal{G}}, y \in \mathcal{Y}_{\mathcal{G}}((u,v))\}$ is the triplet collection of graph $\mathcal{G}$, $s_y(u,v)$ is the score for $\langle u,y,v \rangle$, and $\langle u_t', y, v_t'\rangle$ is one of the $T$ negative triplets sampled from $\mathcal{G}$ by replacing $u$ with $u_t'$ or $v$ with $v_t'$ uniformly:
\scalebox{0.87}{\parbox{1.149\linewidth}{
\begin{align}
    \mathcal{J} &= - \frac{1}{|\mathcal{D}|}\sum_{\langle u,y,v \rangle \in \mathcal{D}}\Bigl(\log \sigma(s_y(\bm{h}_{u}, \bm{h}_{v})) \nonumber \\
    &\qquad \qquad \ \  - \frac{1}{T}{\sum_{t=1}^{T}}{\log (1 - \sigma(s_y(\bm{h}_{u'_t}, \bm{h}_{v'_t})))}\Bigr). \label{eq:obj_unc}
\end{align}
}}
We report the results of KG embedding models, proximity-preserving based embedding methods, graph convolutional networks, and graph attention networks for comparison.
\reviseyq{We} use the same parameter setting as \citeauthor{yang2020heterogeneous}~(\citeyear{yang2020heterogeneous}).

\subsubsection{Datasets}
\begin{table}[!t]
    \footnotesize
    \vspace{-0.05in}
    \centering
    \setlength\tabcolsep{4pt}
    \resizebox{0.96\linewidth}{!}{%
    \begin{tabular}{c|c|c|c|c|c}
    \toprule
        Dataset & $|\mathcal{V}_{\mathcal{G}}|$ & $|\mathcal{E}_{\mathcal{G}}|$ & $|\mathcal{Y}_{\mathcal{G}}|$ & \#Label type & \#Labeled node \\
        \toprule
        PubMed & 63,109 & 244,986    & 10 & 8  & 454 \\
        Yelp   & 82,465 & 30,542,675 & 4  & 16 & 7,417 \\
        \bottomrule
    \end{tabular}
    }
    \vspace{-0.1in}
    \caption{Statistics of two real-life heterogeneous networks on unattributed unsupervised node classification.}
    \label{table:stat_unc}
    \vspace{-0.2in}
\end{table}
\citeauthor{yang2020heterogeneous}~(\citeyear{yang2020heterogeneous}) collected and processed two heterogeneous networks to evaluate graph embedding algorithms. 
PubMed is a biomedical network constructed by text mining and manual processing where nodes are labeled as one of eight types of diseases; Yelp is a business network where nodes may have multiple labels (businesses, users, locations, and reviews).
Statistics are summarized in Table~\ref{table:stat_unc}.


\subsubsection{Results}
In Table~\ref{table:nodeclf}, we observe low F1 scores on both datasets and the difficulty of this task.
Traditional KG embedding methods perform very similarly, but graph neural networks vary dramatically.
RGCN and RGIN adapt the same relational transformations, but RGIN surpasses RGCN because of sum aggregation and MLPs.
HAN and MAGNN explicitly learn the node representations from meta-paths and meta-path neighbors, but these models are evidently easy to overfit to training data because they predict the connectivity with the leaky edge type information.
On the contrary, CompGCN and HGT obtain better scores since CompGCN incorporates semantics by node-relation composition, and HGT captures semantic relations and injects edge dependencies by relation-specific matrices.
Our DMPNN outperforms all baselines by asynchronously learning node embeddings and edge representations in the same aligned space.
Even for the challenging 16-way multi-label classification, DMPNN also works without any node attributes.

\begin{table}[!t]
    \footnotesize
    \vspace{-0.05in}
    \centering
    \setlength\tabcolsep{1.2pt}
    \resizebox{\linewidth}{!}{%
    \begin{tabular}{c|cc|cc}
    \toprule
        \multirow{2}{*}{Models} & \multicolumn{2}{c|}{PubMed} & \multicolumn{2}{c}{Yelp} \\
        & Macro-F1 & Micro-F1 & Macro-F1 & Micro-F1 \\
        \midrule
        \parbox[t]{140pt}{TransE$\ddag$~\tiny{\cite{bordes2013translating}}}
        & \normalsize{11.40} & \normalsize{15.16} & \normalsize{5.05} & \normalsize{23.03} \\
        \parbox[t]{140pt}{DistMult$\ddag$~\tiny{\cite{embedding2015yang}}}
        & \normalsize{11.27} & \normalsize{15.79} & \normalsize{5.04} & \normalsize{23.00}  \\
        \parbox[t]{140pt}{ConvE$\ddag$~\tiny{\cite{dettmers2018convolutional}}}
        & \normalsize{13.00} & \normalsize{14.49} & \normalsize{5.09} & \normalsize{23.02}  \\
        \hline
        \parbox[t]{140pt}{metapath2vec$\ddag$~\tiny{\cite{dong2017metapath2vec}}}
        & \normalsize{12.90} & \normalsize{15.51} & \normalsize{5.16} & \normalsize{23.32} \\
        \parbox[t]{140pt}{HIN2vec$\ddag$~\tiny{\cite{fu2017hin2vec}}}
        & \normalsize{10.93} & \normalsize{15.31} & \normalsize{5.12} & \normalsize{23.25} \\
        \parbox[t]{140pt}{HEER$\ddag$~\tiny{\cite{shi2018easing}}}
        & \normalsize{11.73} & \normalsize{15.29} & \normalsize{5.03} & \normalsize{22.92} \\
        \hline
        \parbox[t]{140pt}{RGCN$\ddag$~\tiny{\cite{schlichtkrull2018modeling}}}
        & \normalsize{10.75} & \normalsize{12.73} & \normalsize{5.10} & \normalsize{23.24} \\
        \parbox[t]{140pt}{RGIN~\tiny{\cite{liu2020neural}}}
        & \normalsize{12.22} & \normalsize{15.41} & \normalsize{5.14}  & \normalsize{23.82} \\
        \parbox[t]{140pt}{CompGCN~\tiny{\cite{vashishth2020composition}}}
        & \normalsize{13.89} & \normalsize{21.13} & \normalsize{5.09} & \normalsize{23.96}\\
        \hline
        \parbox[t]{140pt}{HAN$\ddag$~\tiny{\cite{wang2019heterogeneous}}}
        & \normalsize{9.54} & \normalsize{12.18} & \normalsize{5.10} & \normalsize{23.24} \\
        \parbox[t]{140pt}{MAGNN$\ddag$~\tiny{\cite{fu2020magnn}}}
        & \normalsize{10.30} & \normalsize{12.60} & \normalsize{5.10} & \normalsize{23.24} \\
        \parbox[t]{140pt}{HGT$\ddag$~\tiny{\cite{hu2020heterogeneous}}}
        & \normalsize{11.24} & \normalsize{18.72} & \normalsize{5.07} & \normalsize{23.12} \\
        \hline
        \parbox[t]{140pt}{DMPNN}
        & \bf \normalsize{16.54} & \bf \normalsize{23.13} & \bf \normalsize{12.74} & \bf \normalsize{29.12} \\
        \bottomrule
    \end{tabular}
    }
    \vspace{-0.1in}
    \caption{F1 scores (\%) on unattributed unsupervised node classification. Results of $\ddag$ are taken from ~\cite{yang2020heterogeneous}.}
    \label{table:nodeclf}
    \vspace{-0.13in}
\end{table}
\section{Related Work}
The isomorphism search aims to find all bijections between two graphs.
The subgraph isomorphism search is more challenging, and it has been proven to be an NP-complete problem.
Most subgraph isomorphism algorithms are based on backtracking or graph-index~\cite{ullmann1976an, he2008graphs}.
However, these algorithms are hard to be applied to complex patterns and large data graphs.
The search space of backtracking methods grows exponentially, and the latter requires a large quantity of disk space to index.
Some methods introduce weak rules to reduce search space in most cases, such as candidate region filtering, partial matching enumeration, and ordering~\cite{carletti2018challenging}.
On the other hand, there are many approximate techniques for subgraph counting, such as path sampling~\cite{jha2015path} and color coding~\cite{bressan2019bressan}.
But most approaches are hard to generalize to complex heterogeneous multi-graphs~\cite{sun2020in}.

In recent years, graph neural networks (GNNs) and message passing networks (MPNNs) have achieved success in graph data modeling.
There are also some discussions about isomorphisms.
\citeauthor{xu2019how}~(\citeyear{xu2019how}) and \citeauthor{morris2019weisfeiler}~(\citeyear{morris2019weisfeiler}) \reviseyq{showed} that neighborhood-aggregation schemes are as stronger as Weisfeiler-Leman (1-WL) test.
\revisexin{\citeauthor{chen2020can}~(\citeyear{chen2020can}) \reviseyq{proved} that $k$-WL cannot count all patterns more than $k$ nodes accurately, but the bound of $T$ iterations of $k$-WL grows quickly to $(k+1)2^T$.}
These conclusions encourage researchers to empower message passing and explore the possibilities of neural subgraph counting.
\revisexin{Empirically,}
\citeauthor{liu2020neural}~(\citeyear{liu2020neural}) \reviseyq{combined} graph encoding and dynamic memory networks to count subgraph isomorphisms in an end-to-end way.
They \reviseyq{showed} the memory with linear-complexity read-write operations can significantly improve all encoding models.
A more challenging problem is subgraph isomorphism matching.
NeuralMatch~\cite{ying2020neural} utilizes neural methods and a voting method to detect subgraph matching.
However, it only returns whether one pattern is included in the data graph instead of specific isomorphisms.
Neural subgraph matching is still under discussion.
\revisexin{Besides, graph learning also applies on maximum common subgraph detection~\cite{bai2021glsearch}, providing another possible solution for isomorohisms.}

\section{Conclusion}
In this paper, we theoretically analyze the connection between the edge-to-vertex transform and the duality of isomorphisms in heterogeneous multi-graphs.
We design dual message passing neural networks (DMPNNs) based on the equivalence of isomorphism searching over original graphs and line graphs.
Experiments on subgraph isomorphism counting and matching as well as unsupervised node classification support our theoretical exposition and demonstrate effectiveness.
We also see huge performance boost in small patterns by stacking dual message passing and local relational pooling.
We defer a better integration as future work.
\clearpage
\section{Acknowledgements}
The authors of this paper were supported by the NSFC Fund (U20B2053) from the NSFC of China, the RIF (R6020-19 and R6021-20) and the GRF (16211520) from RGC of Hong Kong, the MHKJFS (MHP/001/19) from ITC of Hong Kong with special thanks to HKMAAC and CUSBLT, and  the Jiangsu Province Science and Technology Collaboration Fund (BZ2021065).
We thank Dr. Xin Jiang for his valuable comments and the Gift Fund from Huawei Noah’s Ark Lab.

\bibliography{aaai22}
\clearpage
\appendix

\section{Appendix A \  Proof of Proposition~\ref{propo:line_iso}}
\begin{proof}
Assume the line graph $\mathcal{H}_1$ is transformed from $\mathcal{G}_1$ by $g_1$ and the line graph $\mathcal{H}_2$ is transformed from $\mathcal{G}_2$ by $g_2$, then
\scalebox{0.85}{\parbox{1.176\linewidth}{
\begin{itemize}
    \item $\forall e=(u,v) \in \mathcal{E}_{\mathcal{G}_1}$, $\mathcal{Y}_{\mathcal{G}_1}((u, v)) =
    \mathcal{X}_{\mathcal{H}_1}(g_1(e))$
    \item $\forall e'=(u',v') \in \mathcal{E}_{\mathcal{G}_2}$, $\mathcal{Y}_{\mathcal{G}_2}((u', v')) =
    \mathcal{X}_{\mathcal{H}_2}(g_2(e'))$.
\end{itemize}
}}
Moreover, based on the isomorphism $f$, we get \\\noindent
\scalebox{0.85}{\parbox{1.176\linewidth}{
\begin{itemize}
    \item $\forall e=(u, v) \in \mathcal{E}_{\mathcal{G}_1}, 
    \mathcal{X}_{\mathcal{H}_1}(g_1((u, v))) = \mathcal{Y}_{\mathcal{G}_1}((u, v)) = \mathcal{Y}_{\mathcal{G}_2}((f(u), f(v))) = \mathcal{X}_{\mathcal{H}_2}(g_2(f(u), f(v)))$,
    \item $\forall e'=(u', v') \in \mathcal{E}_{\mathcal{G}_2},
    \mathcal{X}_{\mathcal{H}_2}(g_2((u', v'))) = \mathcal{Y}_{\mathcal{G}_2}((u', v')) = \mathcal{Y}_{\mathcal{G}_1}((f^{-1}(u'), f^{-1}(v'))) = \mathcal{X}_{\mathcal{H}_1}(g_1(f^{-1}(u'), f^{-1}(v')))$.
\end{itemize}
}}
We find the bijection mapping $g_1((u, v))$ to $g_2((f(u), f(v))$ for any $(u, v) \in \mathcal{E}_{\mathcal{G}_1}$, which is a bijection from $\mathcal{V}_{{\mathcal{H}_1}}$ to $\mathcal{V}_{{\mathcal{H}_2}}$.

Similarly, from the two necessary conditions of isomorphism $f$ about $\mathcal{Y}_{H_1}$ with $\mathcal{X}_{\mathcal{G}_1}$ and  $\mathcal{\mathcal{Y}}_{H_2}$ with $\mathcal{X}_{\mathcal{G}_2}$ and the definition $\forall v \in \mathcal{V}_{\mathcal{G}_1} \mathcal{X}_{G_{1}}(v) = \mathcal{X}_{G_{2}}(f(v))$, we have \\\noindent
\scalebox{0.85}{\parbox{1.176\linewidth}{
\begin{itemize}
    \item $\forall d=g_1^{-1}((u, v)) \in \mathcal{V}_{\mathcal{H}_1}, e=g_1^{-1}((v, w)) \in \mathcal{V}_{\mathcal{H}_1},$ \\
    $\mathcal{Y}_{\mathcal{H}_1}((d, e)) = \mathcal{X}_{\mathcal{G}_1}(v) = \mathcal{X}_{\mathcal{G}_2}(f(v)) = \mathcal{Y}_{\mathcal{H}_2}(g_2^{-1}(f(v)))$,
    \item $\forall d'=g_2^{-1}((u', v')) \in \mathcal{V}_{\mathcal{H}_2}, e'=g_2^{-1}((v', w')) \in \mathcal{V}_{\mathcal{H}_2},$ \\
    $\mathcal{Y}_{\mathcal{H}_2}((d', e')) = \mathcal{X}_{\mathcal{G}_2}(v') = \mathcal{X}_{\mathcal{G}_1}(f^{-1}(v')) = \mathcal{Y}_{\mathcal{H}_1}(g_1^{-1}(f^{-1}(v')))$.
\end{itemize}
}}
Therefore, we conclude that
the two line graphs $\mathcal{H}_1$ and $\mathcal{H}_2$ are isomorphic, where the dual isomorphism $f'$ satisfies $\forall v \in \mathcal{V}_{\mathcal{H}_1}$, $f'(v) = g_2((f(g_1^{-1}(v).\text{source}), f(g_1^{-1}(v).\text{target})))$.
We denote $\mathcal{G}_1 \simeq \mathcal{G}_2 \rightarrow L(\mathcal{G}_1) \simeq L(\mathcal{G}_2)$.
\end{proof}

\begin{figure*}[!htbp]
    \centering
    \includegraphics[width=1.0\textwidth]{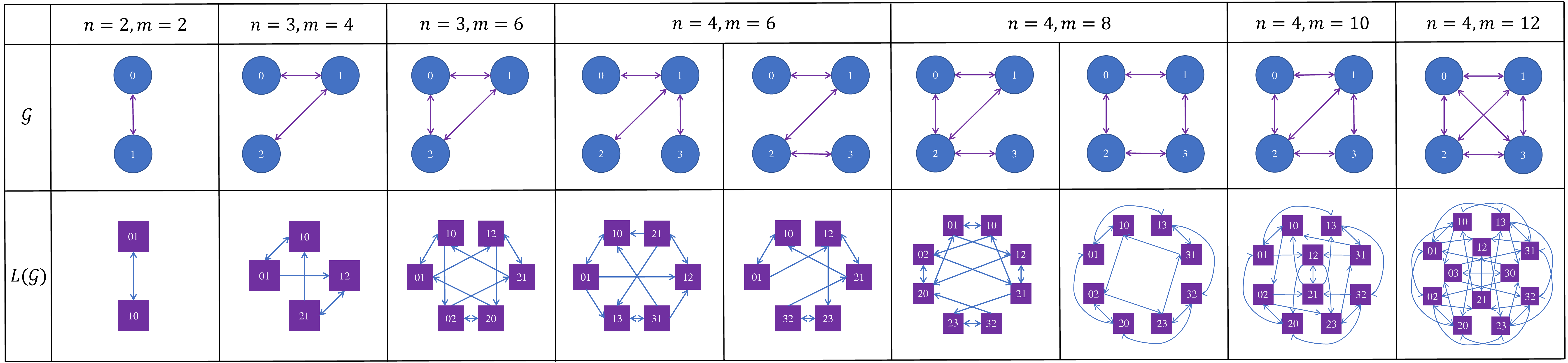}
    \vspace{-0.2in}
    \caption{Simple directed unlabeled graphs with reversed edges and no more than four nodes, and their corresponding line graphs.
    We observe that the number of isomorphisms of two graphs equals to the number of isomorphisms of their line graphs.}
    \label{fig:homo_line}
    \vspace{-0.1in}
\end{figure*}

\section{Appendix B \  Proof of Theorem~\ref{theorem:ext_whitney_iso}}
\label{proof:ext_whitney_iso}
\begin{proof}
{
Assume $\mathcal{G}_1$ and $\mathcal{G}_2$ are two connected directed heterogeneous multigraphs with reversed edges and their isomorphisms are $\mathcal{F} = \{f_1, f_2, \cdots, f_p\}$, 
$\mathcal{H}_1$ and $\mathcal{H}_2$ are their line graphs with isomorphisms $\mathcal{F}' = \{f_1', f_2', \cdots, f_q'\}$,
Let $n_1=|\mathcal{V}_{\mathcal{G}_1}|$, $n_2=|\mathcal{V}_{\mathcal{G}_2}|$, $2m_1=|\mathcal{E}_{\mathcal{G}_1}|=|\mathcal{V}_{\mathcal{H}_1}|$, $2m_2=|\mathcal{E}_{\mathcal{G}_2}|=|\mathcal{V}_{\mathcal{H}_2}|$. To prove Theorem~\ref{theorem:ext_whitney_iso}, we show $|\mathcal{F}| = |\mathcal{F}'|$.

The first step is to prove $|\mathcal{F}| > 0$ is  equivalent to $|\mathcal{F}'| > 0$.
The necessary conditions of $|\mathcal{F}| > 0$ are $|\mathcal{X}_{\mathcal{G}_1}| = |\mathcal{X}_{\mathcal{G}_2}|$ and $|\mathcal{Y}_{\mathcal{G}_1}|=|\mathcal{Y}_{\mathcal{G}_2}|$; the necessary conditions of $|\mathcal{F}'| > 0$ are $|\mathcal{X}_{\mathcal{H}_1}| = |\mathcal{X}_{\mathcal{H}_2}|$ and $|\mathcal{Y}_{\mathcal{H}_1}| = |\mathcal{Y}_{\mathcal{H}_2}|$.
We know that $\mathcal{X}_{\mathcal{H}_1} = \mathcal{Y}_{\mathcal{G}_1} - \phi$ and $\mathcal{X}_{\mathcal{H}_2} = \mathcal{Y}_{\mathcal{G}_2} - \phi$ by the first and second conclusions of Definition~\ref{def:line}, which results in $|\mathcal{Y}_{\mathcal{G}_1}| = |\mathcal{Y}_{\mathcal{G}_2}| \iff |\mathcal{X}_{\mathcal{H}_1}| = |\mathcal{X}_{\mathcal{H}_2}|$.
If there exists one vertex $v \in \mathcal{V}_{\mathcal{G}_1}$ such that $x=\mathcal{X}_{\mathcal{G}_1}(v) \wedge x \notin \mathcal{Y}_{\mathcal{H}_1}$, then there are no two edges $d,e \in \mathcal{E}_{\mathcal{G}_1}$ such that $d.\text{target} = e.\text{source} = v$ by the third conclusion of Definition~\ref{def:line}, which is contradictory to the reversed edge setting.
Hence, $\forall x \in \mathcal{X}_{\mathcal{G}_1}(v), x \in \mathcal{Y}_{\mathcal{H}_1}$, which implies $|\mathcal{Y}_{\mathcal{H}_1}| \geq |\mathcal{X}_{\mathcal{G}_1}|$.
Because elements of $\mathcal{Y}_{\mathcal{H}_1}$ come from $\mathcal{X}_{\mathcal{G}_1}$, we have $|\mathcal{Y}_{\mathcal{H}_1}| \leq |\mathcal{X}_{\mathcal{G}_1}|$.
Therefore, $|\mathcal{Y}_{\mathcal{H}_1}| \ge |\mathcal{X}_{\mathcal{G}_1}| \wedge |\mathcal{Y}_{\mathcal{H}_1}| \le |\mathcal{X}_{\mathcal{G}_1}| \wedge |\mathcal{X}_{\mathcal{G}_1}| = |\mathcal{X}_{\mathcal{G}_2}| \iff |\mathcal{Y}_{\mathcal{H}_1}| = |\mathcal{Y}_{\mathcal{H}_2}|$.
Above all, $|\mathcal{F}| > 0 \iff |\mathcal{F}'| > 0$, which also implies $|\mathcal{F}| = 0 \iff |\mathcal{F}'| = 0$.

The second step is for the proof of $|\mathcal{F}| = |\mathcal{F}'|$ if $|\mathcal{F}| > 0$ when $\mathcal{G}_1$ and $\mathcal{G}_2$ have no self-loop.
By the Definition~\ref{def:line}, we have $n_1=n_2, m_1=m_2$, and there are six scenarios:
\begin{enumerate}[label=\Roman*]
    \item $|\mathcal{X}_{\mathcal{G}_1}| = 0 \wedge |\mathcal{Y}_{\mathcal{G}_1}| = 0$: \\
    Both the graph and the line graph are empty so that both $\mathcal{F}$ and $\mathcal{F}'$ contain the only isomorphism $f'$ is $\{\} \rightarrow \{\}$. Therefore, $|\mathcal{F}| = |\mathcal{F}'|$ holds for $|\mathcal{X}_{\mathcal{G}_1}| = 0 \wedge |\mathcal{Y}_{\mathcal{G}_1}| = 0$.
    \item $|\mathcal{X}_{\mathcal{G}_1}| = 1 \wedge |\mathcal{Y}_{\mathcal{G}_1}| = 0$: \\
    Each connected graph contains one vertex without any edge, and the two nodes must have the same vertex label based on the assumption $\mathcal{F} > 0$.
    Obviously, $|\mathcal{F}|=1$. 
    Their line graphs are empty so that $|\mathcal{F}'|=1$ and the only isomorphism $f'$ is $\{\} \rightarrow \{\}$. Therefore, $|\mathcal{F}| = |\mathcal{F}'|$ holds for $|\mathcal{X}_{\mathcal{G}_1}| = 1 \wedge |\mathcal{Y}_{\mathcal{G}_1}| = 0$.
    \item $|\mathcal{X}_{\mathcal{G}_1}| = 1 \wedge |\mathcal{Y}_{\mathcal{G}_1}| = 1$: \\
    Graphs are able to be regarded as two simple undirected graphs with $n_1 > 1$.
    Theorem~\ref{theorem:whitney_iso} tells us $|\mathcal{F}| = |\mathcal{F}'|$ when $n_1 > 4$.
    Figure~\ref{fig:homo_line} shows all graphs and their line graphs with $1 < n_1 \leq 4$, and we find $|\mathcal{F}| = |\mathcal{F}'|$.
    Specifically, $\bm{C}= \left[\begin{smallmatrix}
2 & 0 & 0 & 0 & 0 & 0 & 0 & 0 & 0 \\
0 & 2 & 0 & 0 & 0 & 0 & 0 & 0 & 0 \\
0 & 0 & 6 & 0 & 0 & 0 & 0 & 0 & 0 \\
0 & 0 & 0 & 6 & 0 & 0 & 0 & 0 & 0 \\
0 & 0 & 0 & 0 & 2 & 0 & 0 & 0 & 0 \\
0 & 0 & 0 & 0 & 0 & 2 & 0 & 0 & 0 \\
0 & 0 & 0 & 0 & 0 & 0 & 8 & 0 & 0 \\
0 & 0 & 0 & 0 & 0 & 0 & 0 & 4 & 0 \\
0 & 0 & 0 & 0 & 0 & 0 & 0 & 0 & 24 
\end{smallmatrix}\right]$ and $\bm{C}'=\left[\begin{smallmatrix}
2 & 0 & 0 & 0 & 0 & 0 & 0 & 0 & 0 \\
0 & 2 & 0 & 0 & 0 & 0 & 0 & 0 & 0 \\
0 & 0 & 6 & 0 & 0 & 0 & 0 & 0 & 0 \\
0 & 0 & 0 & 6 & 0 & 0 & 0 & 0 & 0 \\
0 & 0 & 0 & 0 & 2 & 0 & 0 & 0 & 0 \\
0 & 0 & 0 & 0 & 0 & 2 & 0 & 0 & 0 \\
0 & 0 & 0 & 0 & 0 & 0 & 8 & 0 & 0 \\
0 & 0 & 0 & 0 & 0 & 0 & 0 & 4 & 0 \\
0 & 0 & 0 & 0 & 0 & 0 & 0 & 0 & 24 
\end{smallmatrix}\right]$, where the element $c_{i,j}$ of $\bm{C}$ corresponds to the number of isomorphisms between the $i$-th and the $j$-th graphs, and the element $c'_{i,j}$ of $\bm{C}'$ corresponds to the number of isomorphisms between the $i$-th and the $j$-th line graphs.
    Therefore, $|\mathcal{F}| = |\mathcal{F}'|$ holds for $|\mathcal{X}_{\mathcal{G}_1}| = 1 \wedge |\mathcal{Y}_{\mathcal{G}_1}| = 1$.
    \item $|\mathcal{X}_{\mathcal{G}_1}| > 1 \wedge |\mathcal{Y}_{\mathcal{G}_1}| = 1$: \\
    Graphs are able to be regarded as two simple undirected graphs with $n_1 \geq |\mathcal{X}_{\mathcal{G}_1}| \geq 2$.
    We denote $\overline{\mathcal{F}}$ as the isomorphism sets for $\overline{\mathcal{G}_1}$ and $\overline{\mathcal{G}_2}$, where $\overline{\mathcal{G}_1}$ is the unlabeled graph of $\mathcal{G}_1$, and $\overline{\mathcal{G}_2}$ is the unlabeled graph of $\mathcal{G}_2$, and $\mathcal{F} \subseteq \overline{\mathcal{F}}$. Similarly, we use $\overline{\mathcal{F}'}$ for the isomorphisms of their line graphs. We get $|\overline{\mathcal{F}}| = |\overline{\mathcal{F}'}|$ from \RNum{3}. Assume the dual of $f \in \overline{\mathcal{F}}$ is $f' \in \overline{\mathcal{F}'}$, then we only need to prove $f \notin \mathcal{F} \iff f' \notin \mathcal{F}'$.
    As $|\mathcal{Y}_{\mathcal{G}_1}| = 1$, $f \notin \mathcal{F} \implies (\exists v \in \mathcal{V}_{\mathcal{G}_1}, \mathcal{X}_{\mathcal{G}_1}(v) \neq \mathcal{X}_{\mathcal{G}_2}(f(v)))$.
    That is,
    $\exists v \in \mathcal{V}_{\mathcal{G}_1}, f(v) \in \mathcal{V}_{\mathcal{G}_2} \ \forall d,e \in \mathcal{V}_{\mathcal{H}_1}, d'=f'(d), e'=f'(e) \in \mathcal{V}_{\mathcal{H}_2}, g_1^{-1}(d).\text{target}=g_1^{-1}(e).\text{source} = v \implies \mathcal{Y}_{\mathcal{H}_1}((d, e)) = \mathcal{X}_{\mathcal{G}_1}(v) \neq  \mathcal{X}_{\mathcal{G}_2}(v') = \mathcal{Y}_{\mathcal{H}_2}((d', e'))$, which means $f'$ is also not an isomorphism from $\mathcal{H}_1$ to $\mathcal{H}_2$ so that $f \notin \mathcal{F} \implies f' \notin \mathcal{F}'$. We can also prove $f' \notin \mathcal{F}' \implies f \notin \mathcal{F}$ similarly.
    Therefore, $|\mathcal{F}| = |\mathcal{F}'|$ holds for $|\mathcal{X}_{\mathcal{G}_1}| > 1 \wedge |\mathcal{Y}_{\mathcal{G}_1}| = 1$.
    \item $|\mathcal{X}_{\mathcal{G}_1}| = 1 \wedge |\mathcal{Y}_{\mathcal{G}_1}| > 1$: \\
    Graphs are connected so that $2m_1 \geq |\mathcal{Y}_{\mathcal{G}_1}| > 1$ and $n_1 \geq 2$.
    We use the same notations $\overline{\mathcal{F}}$ and $\overline{\mathcal{F}'}$ in \RNum{4} to prove $f \notin \mathcal{F} \iff f' \notin \mathcal{F}'$.
    As $|\mathcal{X}_{\mathcal{G}_1}| = 1$, $f \notin \mathcal{F} \implies (\exists (u, v) \in \mathcal{E}_{\mathcal{G}_1}, \mathcal{Y}_{\mathcal{G}_1}((u,v)) \neq \mathcal{Y}_{\mathcal{G}_2}((f(u),f(v))))$.
    That is, $\exists (u,v) \in \mathcal{E}_{\mathcal{G}_1}, u'=f(u), v'=f(v), (u',v') \in \mathcal{E}_{\mathcal{G}_2}, e=g_1((u,v)) \in \mathcal{H}_1, e'=g_2((u',v')) \in \mathcal{H}_2, \mathcal{X}_{\mathcal{H}_1}(e) = \mathcal{Y}_{\mathcal{G}_1}((u,v)) \neq \mathcal{Y}_{\mathcal{G}_2}((u',v')) = \mathcal{X}_{\mathcal{H}_2}(e')$,. which means $f'$ is also not an isomorphism from $\mathcal{H}_1$ to $\mathcal{H}_2$ so that $f \notin \mathcal{F} \implies f' \notin \mathcal{F}'$. We can also prove $f' \notin \mathcal{F}' \implies f \notin \mathcal{F}$ similarly.
    Therefore, $|\mathcal{F}| = |\mathcal{F}'|$ holds for $|\mathcal{X}_{\mathcal{G}_1}| = 1 \wedge |\mathcal{Y}_{\mathcal{G}_1}| > 1$.
    \item $|\mathcal{X}_{\mathcal{G}_1}| > 1 \wedge |\mathcal{Y}_{\mathcal{G}_1}| > 1$: \\
    We prove $|\mathcal{F}| = |\mathcal{F}'|$ by $f \notin \mathcal{F} \iff f' \notin \mathcal{F}'$.
    Now $f \notin \mathcal{F} \implies (\exists v \in \mathcal{V}_{\mathcal{G}_1}, \mathcal{X}_{\mathcal{G}_1}(v) \neq \mathcal{X}_{\mathcal{G}_2}(f(v))) \vee (\exists (u, v) \in \mathcal{E}_{\mathcal{G}_1}, \mathcal{Y}_{\mathcal{G}_1}((u,v)) \neq \mathcal{Y}_{\mathcal{G}_2}((f(u),f(v))))$.
    The proof of the two clauses are shown in \RNum{4} and \RNum{5} in several. So we get $f \notin \mathcal{F} \implies f' \notin \mathcal{F}'$ and $f' \notin \mathcal{F}' \implies f \notin \mathcal{F}$, which indicates $|\mathcal{F}| = |\mathcal{F}'|$ holds for $|\mathcal{X}_{\mathcal{G}_1}| > 1 \wedge |\mathcal{Y}_{\mathcal{G}_1}| > 1$.
\end{enumerate}

The third step is to verify that self-loops do not affect the one-to-one property.
Let $\hat{\mathcal{F}}$ to be the isomorphism set for $\hat{\mathcal{G}}_1$ and $\hat{\mathcal{G}}_2$ and $\hat{\mathcal{F}}'$ to be the set for their line graphs $\hat{\mathcal{H}}_1$ and $\hat{\mathcal{H}}_2$, where $\hat{\mathcal{G}}_1$ and $\hat{\mathcal{G}}_2$ are obtained by removing self-loops from $\mathcal{G}_1$ and $\mathcal{G}_2$.
That is,
$\mathcal{V}_{\hat{\mathcal{G}}_1} = \mathcal{V}_{\mathcal{G}_1}$ and
$\mathcal{E}_{\hat{\mathcal{G}}_1} = \mathcal{E}_{\mathcal{G}_1} - \{(v,v) \in \mathcal{E}_{\mathcal{G}_1}\}$ for $\hat{\mathcal{G}}_1$;
$\mathcal{V}_{\hat{\mathcal{H}}_1} = \mathcal{V}_{\mathcal{H}_1} - \{e \in \mathcal{E}_{\mathcal{H}_1}| g_1^{-1}(e).\text{source} =  g_1^{-1}(e).\text{target}\}$,
$\mathcal{E}_{\hat{\mathcal{H}}_1} = \mathcal{E}_{\mathcal{H}_1} - \{(d, e) \in \mathcal{E}_{\mathcal{H}_1}| \forall (v,v) \in \mathcal{E}_{\mathcal{G}_1}, g_1^{-1}(d).\text{target}=g_1^{-1}(e).\text{source} = v\}$ for $\hat{\mathcal{H}}_1$.
We have proved there is a one-to-one correspondence between $\hat{\mathcal{F}}$ and $\hat{\mathcal{F}}'$ so that we can denote the dual of $\hat{f} \in \hat{\mathcal{F}}$ is $\hat{f}' \in \hat{\mathcal{F}}'$.
Considering the isomorphism is defined between vertex sets, and removing self-loops does not affect the connectivity, we have $f \in \mathcal{F} \implies f \in \hat{\mathcal{F}}$ and $f' \in \mathcal{F}' \implies f' - \Omega \in \hat{\mathcal{F}}'$, where $\Omega = \{(v, v') | \forall v \in \mathcal{V}_{\mathcal{H}_1} \ \forall v' \in \mathcal{V}_{\mathcal{H}_2}, (g_1^{-1}(v).\text{source} = g_1^{-1}(v).\text{target}) \wedge (g_2^{-1}(v').\text{source} = g_2^{-1}(v').\text{target})\}$ includes all possible mappings from vertices in $\mathcal{H}_1$ that correspond to self-loops in $\mathcal{G}_1$ to vertices in $\mathcal{H}_2$ that correspond to self-loops in $\mathcal{G}_2$.

It is easy to get $|\mathcal{F}'| \leq |\mathcal{F}|$:
assume $f_1',f_2' \in \mathcal{F}'$ and $f_1' \neq f_2'$, then $f_1' - \Omega, f_2' - \Omega \in \hat{\mathcal{F}}'$ and $f_1' - \Omega \neq f_2' - \Omega$, and so there are two different isomorphisms $f_1, f_2 \in \mathcal{F}$ whose dual isomorphisms are $f_1' - \Omega $ and $f_2' - \Omega$ because of $|\hat{\mathcal{F}}| = |\hat{\mathcal{F}}'|$.

We consider two scenarios for the proof of $|\mathcal{F}| \leq |\mathcal{F}'|$:
\begin{enumerate}[label=\Roman*]
    \item if $\hat{f} \notin \mathcal{F}$: \\
    $\hat{f} \notin \mathcal{F}$ indicates that there exists one vertex $v$ of $\mathcal{G}_1$ with self-loops while its corresponding $v'=\hat{f}(v)$ of $\mathcal{G}_2$ has self-loops with different labels or no self-loop, i.e., $\mathcal{Y}_{\mathcal{G}_1}((v,v)) \neq \mathcal{Y}_{\mathcal{G}_2}((v',v'))$.
    We get $\mathcal{X}_{\mathcal{H}_1}(g_1((v,v))) = \mathcal{Y}_{\mathcal{G}_1}((v,v)) \neq \mathcal{Y}_{\mathcal{G}_2}((v', v')) = \mathcal{X}_{\mathcal{H}_2}(g_2((v', v')))$ so that $\forall f' \in \mathcal{F}', f'(g_1((v,v))) \neq g_2((v', v'))$.
    Therefore, $f'$ cannot obtained adding specific mappings to $\hat{f}'$, i.e., $\forall \Gamma \subseteq \Omega, \hat{f}' + \Gamma \notin \mathcal{F}'$.
    \item if $\hat{f} \in \mathcal{F}$: \\
    We prove that there must be one and only one $f' \in \mathcal{F}'$ that satisfies
    $f' = \hat{f}' \cup \Delta$, where $\Delta = \{(v, v') |\forall (u,u) \in \mathcal{E}_{\mathcal{G}_1} \ \forall v \in \mathcal{V}_{\mathcal{H}_1} \ \exists v' \in \mathcal{V}_{\mathcal{H}_2}, (g_1((u,u)) = v) \wedge (g_2((\hat{f}(u), \hat{f}(u))) = v')\}$.
    For the existence, we can check $\hat{f}' \cup \Delta \in \mathcal{F}'$ by Definition~\ref{def:line}.
    For the uniqueness, if $f'_1,f'_2 \in \mathcal{F}'$ such that $f'_1 = \hat{f}' \cup \Delta_1$, $f'_2 = \hat{f}' \cup \Delta_2$, where $\Delta_1 \neq \Delta_2$, then there exists one vertex $v$ of $\mathcal{H}_1$ and two different vertices $v'_1, v'_2$ of $\mathcal{H}_2$ associated with the same vertex labels such that $(v, v'_1) \in \Delta_1, (v, v'_2) \in \Delta_2$, which indicates $v_1'=g_2((\hat{f}(u), \hat{f}(u)))$ and $v_2'=g_2((\hat{f}(u), \hat{f}(u)))$, where $v = g_1((u, u))$.
    It is impossible because $g_1, g_2$, and $\hat{f}$ are one-to-one.
\end{enumerate}
Above all, $|\mathcal{F}| = |\mathcal{F}'|$ for any directed heterogeneous multigraphs with reversed edges, showing that there is a one-to-one correspondence between isomorphisms of the graphs and isomorphisms of their line graphs.
}\label{proof:ext_whitney_iso}
\end{proof}

\section{Appendix C \  Reparameterizations of DMPNNs}

\subsubsection{Filter Decomposition} Compared with simple graphs, directions and edge labels are non-negligible structral properties in directed heterogeneous multigraphs.
We use two groups of parameters to substitute the $\bm \theta_1^{(k)}$ in Eq.~(\ref{eq:hete_node_update}): 
\scalebox{0.90}{\parbox{1.111\linewidth}{
\begin{align}
       & -\frac{\bm \theta_1^{(k)-}}{\lambda_{{\mathcal{G}}{\text{max}}}} \frac{(\bm{\hat{B}}_{\mathcal{G}} - \bm{B}_{\mathcal{G}})}{2} \bm z^{(k-1)}
       +\frac{\bm \theta_1^{(k)+}}{\lambda_{{\mathcal{G}}{\text{max}}}} \frac{(\bm{\hat{B}}_{\mathcal{G}} + \bm{B}_{\mathcal{G}})}{2} \bm z^{(k-1)} \nonumber \\
       = & \frac{\bm \theta_1^{(k)+} - \bm \theta_1^{(k)-}}{2 \lambda_{{\mathcal{G}}{\text{max}}}} \bm{\hat{B}}_{\mathcal{G}}\bm z^{(k-1)} + \frac{\bm \theta_1^{(k)-} + \bm \theta_1^{(k)+}}{2 \lambda_{{\mathcal{G}}{\text{max}}}} \bm{B}_{\mathcal{G}}\bm z^{(k-1)} \nonumber \\
       \mapsto & \frac{\bm \theta_1^{(k)}}{\lambda_{{\mathcal{G}}{\text{max}}}} \bm{B}_{\mathcal{G}} \bm z^{(k-1)}, \nonumber
\end{align}
}}
where the former aggregates the in-edges and the latter accumulates the out-edges.
Similarly, two groups of parameters are involved to handle the incoming and outcoming messages in the line graph: $\bm \gamma^{(k)}_1$ is parameterized as $\bm \gamma^{(k)-}_1$ and $-\bm \gamma^{(k)+}_1$ and $\frac{\bm \gamma_1^{(k)}}{\lambda_{{\mathcal{H}}{\text{max}}}} \bm{\hat{B}}_{\mathcal{G}}^\top \bm h^{(k-1)}$ is decoupled as:
\scalebox{0.90}{\parbox{1.111\linewidth}{
\begin{align}
       & \frac{\bm \gamma_1^{(k)-}}{\lambda_{{\mathcal{H}}{\text{max}}}} \frac{(\bm{\hat{B}}_{\mathcal{G}} - \bm{B}_{\mathcal{G}})^\top}{2} \bm h^{(k-1)}
       -\frac{\bm \gamma_1^{(k)+}}{\lambda_{{\mathcal{H}}{\text{max}}}} \frac{(\bm{\hat{B}}_{\mathcal{G}} + \bm{B}_{\mathcal{G}})^\top}{2} \bm h^{(k-1)} \nonumber \\
       = & \frac{\gamma_1^{(k)-} - \gamma_1^{(k)+}}{2 \lambda_{{\mathcal{H}}{\text{max}}}} \bm{\hat{B}}_{\mathcal{G}}^\top \bm h^{(k-1)} - \frac{\gamma_1^{(k)-} + \gamma_1^{(k)+}}{2 \lambda_{{\mathcal{H}}{\text{max}}}} \bm{B}_{\mathcal{G}}^\top \bm h^{(k-1)} \nonumber \\
       \mapsto & \frac{\bm \gamma_1^{(k)}}{\lambda_{{\mathcal{H}}{\text{max}}}} \bm{\hat{B}}_{\mathcal{G}}^\top \bm h^{(k-1)}, \nonumber
\end{align}
}}
where the former is used for the source and the latter calculates the target part.
We also simplify the symbols by $\bm \theta^{(k)}_0 \mapsto \bm \theta^{(k)}_0 - \bm \theta^{(k)}_1$ and $\bm \gamma^{(k)}_0 \mapsto \bm \gamma^{(k)}_0 - \bm \gamma^{(k)}_1$.

\subsubsection{Filter Reparameterization}
To handle multi-channel features, e.g., properties,  types, and some structures, we use $l^{(0)}$-dim vectors instead of scalars to represent the initial vertex and edge representations, and $l^{(k-1)} \times l^{(k)}$ input-output matrices to serve as filters $\bm \Theta^{(k)}$ and $\bm \Gamma^{(k)}$ for the $k$-th DMPNN.
\revisexin{Specifically, we reparameterize $\bm \theta^{(k)}_0 \bm h^{(k-1)}$ with $\bm{W}^{(k) \top}_{\theta_0} \bm{h^{(k-1)}}$, $\frac{\bm \theta_1^{(k)}}{\lambda_{{\mathcal{G}}{\text{max}}}} \bm{\hat{B}}_{\mathcal{G}}\bm z^{(k-1)}$ with $\bm{\hat{B}}_{\mathcal{G}} (\bm{W}^{(k) \top}_{\theta_1} \bm z^{(k-1)})$, $\bm \gamma^{(k)}_0 \bm z^{(k-1)}$ with $\bm{W}^{(k) \top}_{\gamma_0} \bm{z^{(k-1)}}$, $\frac{\bm \gamma^{(k)}}{\lambda_{{\mathcal{H}}{\text{max}}}} \bm{\hat{B}}_{\mathcal{G}}\bm h^{(k-1)}$ with $\bm{\hat{B}}_{\mathcal{G}} (\bm{W}^{(k) \top}_{\gamma_1} \bm h^{(k-1)})$, and $\frac{\bm \gamma^{(k)}_1}{\lambda_{\mathcal{H}\text{max}}}(\bm{D}_{\mathcal{H}} + \bm{I}_{m}) \bm z^{(k-1)}$ with $(\bm{D}_{\mathcal{H}} + \bm{I}_{m})(\bm{W}^{(k) \top}_{\gamma_1} \bm z^{(k-1)})$.}
\revisexin{Combining with filter decomposition, we get the parallel end-to-end dual message passing:}
\scalebox{0.90}{\parbox{1.111\linewidth}{
\begin{align}
    \bm H^{(k)} &= \bm H^{(k-1)} \bm{W}^{(k)}_{\theta_0} - (\bm{\hat{B}}_{\mathcal{G}} - \bm{B}_{\mathcal{G}}) \bm Z^{(k-1)}\bm{W}^{(k)}_{\theta_1^-} \nonumber \\
    &\qquad\qquad\qquad\quad + (\bm{\hat{B}}_{\mathcal{G}} + \bm{B}_{\mathcal{G}}) \bm Z^{(k-1)}\bm{W}^{(k)}_{\theta_1^+}, \nonumber \\
    \bm Z^{(k)} &= \bm Z^{(k-1)} \bm{W}^{(k)}_{\gamma_0} + 2(\bm{D}_{\mathcal{H}} + \bm{I}_{m}) \bm Z^{(k-1)} (\bm{W}^{(k)}_{\gamma_1^-} - \bm{W}^{(k)}_{\gamma_1^+}) \nonumber \\
    &\qquad\qquad\qquad\quad - (\bm{\hat{B}}_{\mathcal{G}} - \bm{B}_{\mathcal{G}})^\top \bm H^{(k-1)} \bm{W}^{(k)}_{\gamma_1^-} \nonumber \\
    &\qquad\qquad\qquad\quad + (\bm{\hat{B}}_{\mathcal{G}} + \bm{B}_{\mathcal{G}})^\top \bm H^{(k-1)} \bm{W}^{(k)}_{\gamma_1^+}. \nonumber
\end{align}
}}

\section{Appendix D \  Efficiency Comparison}
\revisexin{In a message-passing framework, the computation cost is linear to the number of layers $K$ and edges $|\mathcal{E}|$, i.e., $\mathcal{O}(K |\mathcal{E}|)$. If we simply apply GNNs to the dual graph, the computation cost increases to $\mathcal{O}(K |\mathcal{E}|^2)$. Our DMPNN does not explicitly construct the line graph but model edge representations with the dual graph property, shown in the ``Dual Message Passing Mechanism" subsection. Eq. (5) updates edge representation with the unoriented incidence matrix, which is a sparse matrix with $2 |\mathcal{E}|$ non-zero elements. Thus, the proposed dual message passing is still $\mathcal{O}(K |\mathcal{E}|)$. Even adding reversed edges, the additional cost is still acceptable. We list the evaluation time in the following Table~\ref{table:time}.
We can see that message passing and dual message passing are still efficient.}

\begin{table}[h]
    \footnotesize
    \centering
    \setlength\tabcolsep{3pt}
    \resizebox{\linewidth}{!}{%
    \begin{tabular}{rl|c|c|c|c}
    \toprule
        \multicolumn{2}{c|}{Models} & Erd\H{o}s-Renyi & Regular & Complex & MUTAG \\
        \midrule
        \parbox[t]{50pt}{\multirow{2}{*}{CNN}}
        & w/o rev & 10.0 & 9.95 & 36.55 & 1.18 \\
        & w/ rev  & 10.1 & 9.98 & 37.27 & 1.19 \\
        \hline
        \parbox[t]{50pt}{\multirow{2}{*}{LSTM}}
        & w/o rev & 10.26 & 10.24 & 38.16 & 1.32 \\
        & w/ rev  & 10.36 & 10.27 & 40.25 & 1.30 \\
        \hline
        \parbox[t]{50pt}{\multirow{2}{*}{TXL}}
        & w/o rev & 10.56 & 10.37 & 39.17 & 1.23 \\
        & w/ rev  & 10.74 & 10.46 & 41.52 & 1.51 \\
        \hline
        \parbox[t]{50pt}{\multirow{2}{*}{RGCN}}
        & w/o rev & 2.69 & 2.71 & 7.98 & 0.69 \\
        & w/ rev  & 2.84 & 2.90 & 9.21 & 0.72 \\
        \hline
        \parbox[t]{50pt}{\multirow{2}{*}{RGIN}}
        & w/o rev & 2.83 & 2.83 & 8.14 & 0.70 \\
        & w/ rev  & 2.89 & 2.93 & 9.30 & 0.77 \\
        \hline
        \parbox[t]{50pt}{\multirow{2}{*}{CompGCN}}
        & w/o rev & 6.76 & 5.51 & 31.32 & 1.22 \\
        & w/ rev  & 6.79 & 5.56 & 31.99 & 1.55 \\
        \hline
        \parbox[t]{50pt}{\multirow{2}{*}{\bf DMPNN}}
        & w/o rev & 6.58 & 5.14 & 29.55 & 1.16 \\
        & w/ rev  & 6.73 & 5.38 & 30.28 & 1.44 \\
        \hline
        \hline
        \parbox[t]{50pt}{\multirow{2}{*}{Deep-LRP}}
        & w/o rev & 7.81 & 7.94 & 34.12 & 1.25 \\
        & w/ rev  & 7.96 & 9.22 & 60.64 & 1.46 \\
        \hline
        \parbox[t]{50pt}{\multirow{2}{*}{\bf DMPNN-LRP}}
        & w/o rev & 9.44 & 8.14 & 35.19 & 1.29 \\
        & w/ rev  & 9.75 & 11.07 & 62.05 & 1.63 \\
        \bottomrule
    \end{tabular}
    }
    \vspace{-0.1in}
    \caption{Average time (in seconds) on test data in 100 runs. Time of batchifying is excluded, but mapping integers to vectors by MultiHot is included.}
    \label{table:time}
    \vspace{-0.2in}
\end{table}

\end{document}